# Large Language Models in Mental Health: A Systematic Review of Applications, Innovations, and Ethical Challenges

Yisong Chen[a*], Yifan Gao[b], Sijing Yu[c], Chuqing Zhao[d] and Yang Lu[e]

*[a] College of Computing, Georgia Institute of Technology, Atlanta, US; [b] Department of Information System and Cyber Security, The University of Texas at San Antonio, San Antonio, US; [c] Department of Computer Science and Engineering, Texas A&M University, US; Harvard University, School of Engineering and Applied Sciences, US; [e] School of Computing and Artificial Intelligence, Beijing Technology and Business School, Beijing, China*

[*]*yisongchen@gatech.edu*

# Large Language Models in Mental Health: A Systematic Review of Applications, Innovations, and Ethical Challenges

We present a review on the applications of large language models (LLMs) in health, e.g., social media analysis, clinical conversational agents, therapy support tools, prompt engineering, multimodal learning, and ethical considerations. We integrate findings from interdisciplinary studies utilizing diverse data sources such as social media posts, electronic medical records, and multimodal inputs to enable early detection of depression, suicide risk assessment, personalized therapy support, and psychoeducational content generation. Our review highlights advancements in LLM models and annotation strategies that enhance interpretability and clinical relevance, while we also emphasize the critical role of prompt engineering for domain adaptation. We also discuss emerging multimodal fusion techniques integrating text, speech, and sensor data for improved mental health diagnosis and monitoring. Finally, we address ongoing ethical, sociotechnical, and regulatory challenges, and advocate frameworks to ensure safe, equitable, and accountable deployment of LLMs in real-world mental health care.

## 1. Introduction

Depression, anxiety, and suicidal behavior together account for a substantial share of the global disease burden. Yet access to timely diagnosis and care remains uneven, constrained by social stigma, financial barriers, and a shortage of mental-health professionals. Meanwhile, social media sites such as Twitter and Reddit have become spaces where people openly describe their mental-health experiences. These self-expressive posts create large, naturally occurring datasets that researchers can analyze to detect and monitor distress signals in real time. Recent research has begun to align social media signals more closely with clinical practice, for example by mapping posts to validated instruments such as the PHQ-9 (Zhang et al. 2023) or by distinguishing levels of depression severity rather than treating detection as a binary task (Zhang, Yang, and Ananiadou 2023). Other directions emphasize explainability to ensure outputs are meaningful to clinicians and patients (Wang, Inkpen, and Kirinde Gamaarachchige 2024), highlight cross-lingual adaptation in diverse contexts such as Chinese social media (Zheng, Guo, and Hong 2024), and extend from individual-level detection to population-level trend discovery through LLM-powered topic modeling (Zhao and Chen 2025).

Despite this momentum, systematic assessments consistently describe the field as nascent. Reviews (Guo et al. 2024c; Ni and Jia 2025; Omar and Levkovich 2025) highlight promising advances in detection, severity assessment, and conversational support, but also point to enduring challenges: reliance on small or imbalanced datasets, limited clinical validation, uncertain generalizability, and unresolved concerns around bias, privacy, and accountability. Against this backdrop, this review analyzes recent applications of large language models (LLMs) in mental health and examines the methodological innovations and ongoing challenges shaping their development. We discuss how LLMs assist in screening, diagnosis, and therapy and outline the ethical, technical, and practical challenges in their application to mental health care.

The rapid proliferation of LLMs has transformed natural language understanding and created new collaborations across computer science, psychiatry, and public health. Built upon earlier advances in natural language processing (NLP), these models can interpret and generate human language with contextual accuracy across diverse sources such as clinical notes and social media. Their ability to combine linguistic, behavioral, and contextual cues at scale opens possibilities for early detection, continuous assessment, and personalized intervention that were not achievable through conventional computational methods. However, this progress also introduces new responsibilities to ensure that models trained on diverse data remain clinically interpretable, culturally inclusive, and ethically sound. This review emphasizes the need for clear evaluation standards and practical frameworks that connect algorithmic advances with real clinical impact.

The paper is structured as follows. Section 2 introduces the background and preliminary concepts and traces the progression from early machine learning and deep learning models to the emergence of Transformers and large language models in mental health research. Section 3 describes the research methodology, details the PRISMA-guided review process, database selection, and criteria for study inclusion and evaluation. Section 4 reviews the main applications of generative AI and LLMs in mental health across social media analysis, clinical conversational agents, and therapy or decision-support tools. Section 5 examines recent innovations and challenges, such as advances in multimodal learning as well as the ethical and sociotechnical considerations surrounding the responsible deployment of LLMs in digital mental health contexts. Section 6 discusses methodological advances, practical implications, and unresolved challenges that shape the translation of LLMs into clinical and public health contexts. Finally, Section 7 concludes the study by summarizing key insights and presenting future directions for the responsible development and deployment of generative AI in mental health.

## 2. Literature Review

Early research on social media–based mental health analysis laid the groundwork for computational approaches to psychological state modeling. These works demonstrated that user-generated text contains linguistic and behavioral signals of depression, anxiety, and related conditions, which can be systematically analyzed for early detection and intervention. Comparative studies revealed key trade-offs between traditional ML and newer DL models. ML algorithms such as logistic regression and random forests offered clear interpretability and feature-importance analysis, while DL architectures such as ALBERT and GRU captured deeper semantic patterns by learning hierarchical representations from text (Ding et al. 2025). Time-aware detection frameworks and feature-engineered models using sentiment lexicons, textual similarity, and network-based attributes further improved the early identification of depressive states (BrilBarniv et al. 2017; Cacheda et al. 2019; Chiong, Budhi, and Dhakal 2021; Lachmar et al. 2017).

With advances in neural networks, deep learning methods became increasingly effective in extracting early depression signals from multimodal social media data. Systems combining textual and visual features demonstrated the promise of multimodal learning for psychological state recognition (Lin et al. 2020). Similarly, studies that used hierarchical neural networks and attention mechanisms improved classification accuracy and interpretability by identifying linguistically salient features in depressive language (Husseini Orabi et al. 2018; Malhotra and Jindal 2022). Beyond binary detection, new approaches began incorporating ordinal classification to capture depression severity levels, reflecting more clinically meaningful distinctions (Naseem et al. 2022).

The emergence of Transformer architectures marked a significant turning point. By introducing self-attention mechanisms, Transformers enabled models to capture long-range dependencies in language and process sequences in parallel, outperforming recurrent architectures in both speed and accuracy. This paradigm shift led to the development of models pre-trained on large corpora and fine-tuned for domain-specific tasks, significantly advancing mental health detection from social media (Dalal, Jain, and Dave 2024; Guo et al. 2024c).

Based on this foundation, LLMs have rapidly transformed the field. Defined as Transformer-based architectures with billions of parameters (Guo et al. 2024c), LLMs extend capabilities to zero-shot and few-shot learning and enable robust performance even in data-scarce mental health contexts. LLMs have been applied to augment training datasets with synthetic examples (Bucur 2024), integrate medical knowledge to enhance explainability (Dalal et al. 2025; Lan et al. 2024), and improve classification across depression, PTSD, and anxiety through context-aware embeddings (Kim, Imieye, and Yin 2025; Shah et al. 2025). Novel methods such as PHQ-aware contrastive learning and sentiment-guided transformers have pushed forward symptom-level and severity-aware classification (Zhang et al. 2023; Zhang, Yang, and Ananiadou 2023), while interpretable LLM frameworks like MentaLLaMA have emphasized human-centered explanations (Yang et al. 2023, 2024).

LLMs have also been explored for high-stakes tasks such as suicidality detection. Studies applied zero-shot prompting to identify suicidal ideation in low-resource settings (Nikmehr et al. 2025), and others used LLMs to extract evidence justifying pre-annotated suicide risk labels (Alhamed, Ive, and Specia 2024b). Broader evaluations benchmarked GPT-3.5 and GPT-4 against domain-specific models for health-related text classification, demonstrating promising results in zero-shot and augmentation settings (Guo et al. 2024b). Additionally, interdisciplinary directions have emerged, including privacy-preserving IoT–LLM frameworks for continuous monitoring (Abdmeziem and Ahmed Nacer 2025) and conversational AI systems that combine LLMswith explainable AI (XAI) to deliver human-readable rationales for depression detection (Belcastro et al. 2025).

Despite these advances, several gaps remain. Many ML and DL studies suffered from small, imbalanced datasets and limited generalizability. While LLM-based systems demonstrate impressive accuracy and flexibility, they face persistent challenges around interpretability, reliability, multimodal integration, and ethical safeguards. Furthermore, clinical translation is still limited, with few studies evaluating these models in real-world healthcare settings. This review addresses these gaps by systematically synthesizing the state of the art in LLM applications for mental health, focusing on social media–based detection of depression and suicidality. This review synthesizes the current state of LLM applications in mental health, with particular attention to social-media-based detection of depression and suicidality. It also identifies research priorities related to clinical utility, privacy, and responsible deployment.

## 3. Research Method

This review follows the PRISMA 2020 (Preferred Reporting Items for Systematic Reviews and Meta-Analyses) guidelines to ensure rigor, transparency, and reproducibility. We conducted a systematic search across seven major scholarly databases: IEEE Xplore, ACM Digital Library, PubMed, Scopus, ScienceDirect, SpringerLink, and Google Scholar. To capture both computational and clinical perspectives, additional manual checks were conducted on OpenReview. Search queries combined the terms "large language models" OR "LLM" OR "transformer" with "mental health", "psychiatry", "psychology", "depression", "anxiety", or "suicidality". Table 1 summarizes the databases and corresponding search schemes.

Table 1. Search Engines and Databases for Manual Search

| Source | Search Scheme |
|---|---|
| **IEEE Xplore** https://ieeexplore.ieee.org/ | Title, Topic, Author, Keywords, Abstract |
| **ACM Digital Library** https://dl.acm.org/ | Title, Keywords |
| **PubMed** https://pubmed.ncbi.nlm.nih.gov/ | Keywords, Full Text |
| **ScienceDirect** https://www.sciencedirect.com/ | Title, Keywords |
| **SpringerLink** https://link.springer.com/ | Title, Keywords |
| **Scopus** https://www.scopus.com/ | Title, Keywords, Abstract |
| **Google Scholar** https://scholar.google.com/ | Topic, Author, Full Text |

The initial search retrieved 304 articles. After removing duplicates, we applied a structured screening process based on the following inclusion and exclusion criteria.

### *3.1 Inclusion Criteria*

- Studies focusing on large language models or Transformer-based architectures in the context of mental health.
- Applications including, but not limited to:
  - Detection of depression, anxiety, suicidality, or related conditions from text or multimodal data.
  - Social media analysis for mental health signals.
  - Clinical conversational agents, therapy support tools, or decision aids.
  - Ethical, regulatory, or sociotechnical considerations of LLM use in mental health.
- Peer-reviewed journal articles, conference proceedings, and systematic reviews.
- Publications in English.
- Studies presenting empirical evaluations, frameworks, case studies, or conceptual models with relevance to mental health practice or research.

### *3.2 Exclusion Criteria*

- Studies without a primary focus on LLMs or Transformer-based models.
- Articles covering general NLP or AI applications with no connection to mental
- health outcomes.
- Non-peer-reviewed publications (e.g., editorials, commentaries, blog posts).
- Duplicate articles or papers unavailable in full text.
- Articles published in languages other than English.

After applying these criteria, 115 papers were shortlisted for full-text evaluation. A second round of screening for methodological transparency, reproducibility, and clinical relevance reduced the pool to a final set of 92 high-quality studies, which form the evidence base for this review.

Figure 1 illustrates the full PRISMA 2020 workflow for search, screening, and selection.

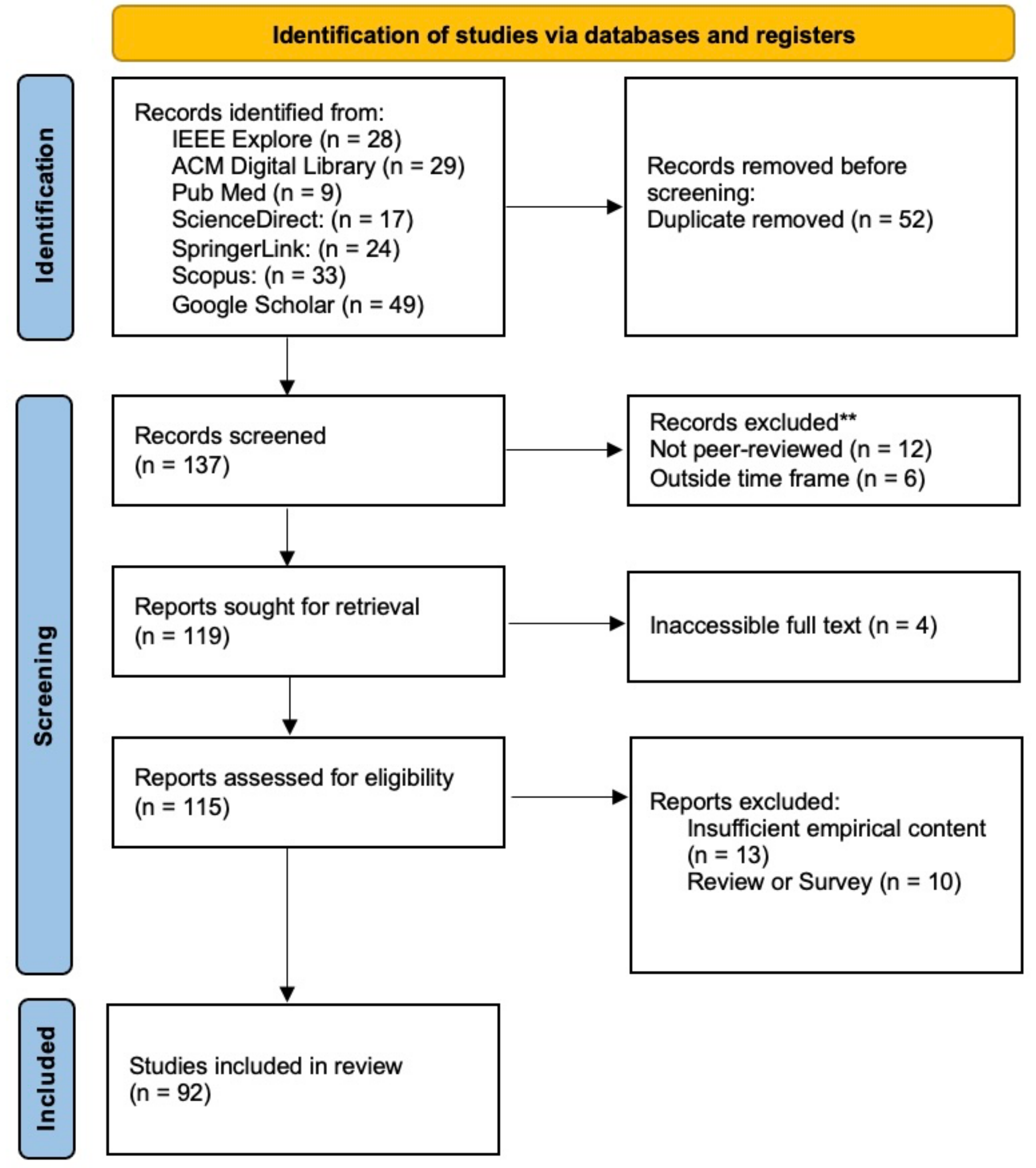

Figure 1. PRISMA Flow Diagram for Study Selection

A structured data extract form was used to standardize information across studies, including model architecture, input modality, mental health domain, data sources, and evaluation metrics. Risk of bias was assessed through criteria such as dataset representativeness, transparency in prompt engineering or fine-tuning, and reproducibility of reported results. Studies were flagged if they lacked clear methodology, relied on proprietary data without explanation, or used performance metrics inconsistently. Data analysis was conducted with Python (Pandas, Seaborn) to generate visualizations of model use cases, performance benchmarks, and domain-specific application trends. The review synthesizes current capabilities, identifies methodological and ethical challenges, and highlights key gaps for future research at the intersection of generative AI and mental health.

### *3.3 Data Usage*

LLM-powered mental health conversational agents rely on diverse, high-quality data sources and efficient knowledge integration techniques to deliver accurate, complete personalized support. As mentioned by (Jin et al. 2025), these data sources include EMR (electronic medical records) (Abbasian et al. 2025), counseling notes (Stade et al. 2024) and more. Details are presented in table 2.

Table 2. Main categories of data sources for LLM-powered mental health conversational agents

| Data Source Category | Description / Example | Related Studies |
|---|---|---|
| **Electronic Medical Records (EMRs)** | Longitudinal clinical data including diagnoses, treatments, and outcomes used for personalized mental health assessment and triage within conversational agents. | Cardamone et al. (2025) |
| **Counseling Notes** | Real or anonymized transcripts and session notes from therapists or behavioral coaches capturing authentic therapeutic dialogue. | Stade et al. (2024) |
| **Social Media Content** | Naturalistic user-generated content (e.g., Twitter, Reddit) expressing mental health experiences, emotions, and symptoms for screening and conversational support agents. | Suri et al. (2023); Xu et al. (2024a) |
| **Simulated and Synthetic Conversations** | Expert-crafted or multi-agent system–generated mental health dialogues allowing controlled evaluation and training of LLMs without personal data risks. | De et al. (2025); Louie et al. (2024); Wang et al. (2024) |
| **Public Clinical and Research Datasets** | Annotated real-world counseling transcripts and GPT-generated sessions from clinical trials used for benchmarking conversational agents. | Singhal et al. (2025); Xu et al. (2024a) |
| **Multimodal and Multilingual Health Data** | Integrations of diverse data types including text, audio, images, and physiological signals for richer agent context and personalization. | Berrezueta et al. (2024) |

## 4. Applications

This section reviews the main applications of generative AI and LLMs in mental health across three areas: social-media analysis, clinical conversational agents, and therapy or decision-support tools. These areas illustrate how LLMs have been applied to meet mental-health needs—through large-scale detection and monitoring of disorders, interactive support for patients and clinicians, and assistance with routine clinical tasks. The following subsections describe how LLMs use diverse data sources, computational methods, and integration frameworks to support new approaches to screening, diagnosis, treatment, and patient engagement.

### *4.1 Social Media Analysis for Depression and Suicidal Ideation*

#### *4.1.1 The Role of Social Media in Mental Health Surveillance*

In the digital era, individuals routinely share personal experiences, emotions, and reflections on social media platforms. This continuous stream of user-generated content (UGC) offers a rich resource to study mental health states at both individual and population levels. Platforms such as Twitter, Reddit, and other online communities have become central sources for researchers studying mental health, as they provide large-scale, naturally occurring data that is difficult to obtain in clinical environments (Cai et al. 2024; Farruque et al. 2024; Hua et al. 2025; Zhang et al. 2023).

These platforms are particularly valuable for capturing unfiltered expressions of psychological distress that may go unreported through traditional healthcare systems. They enable observation of real-time shifts in users' mental states and behaviors, including early indicators of emotional decline such as expressions of hopelessness, withdrawal, or suicidal thoughts. Moreover, social media allows researchers to access narratives from populations that are underserved or hesitant to seek professional care and thereby help fill important gaps in mental health surveillance.

Beyond individual expressions, social media communities foster peer support and informal help-seeking behaviors, and offer insight into the social dynamics of mental health coping strategies. This aspect offers a complementary dimension to clinical datasets and shows how individuals articulate and respond to mental health challenges in collective digital environments.

At the same time, the informal nature of language on these platforms—often marked by slang, abbreviations, and cultural references—presents challenges for computational analysis. Self-disclosure norms vary widely across users and communities, and introduce potential biases related to demographics, platform culture, and topic sensitivity. These factors complicate both data interpretation and model generalizability. Ethical considerations, particularly regarding privacy, consent, and potential harms of passive data collection, will be addressed in Section 5.2.

#### *4.1.2 Data Sources and Annotation Strategies*

Social media platforms offer large-scale, real-world text data that is frequently used to train models for mental health classification. Among the studies reviewed, a majority relied on publicly available user posts, with platform selection varying based on linguistic community, accessibility, and relevance to specific research questions. While platforms such as Twitter and Reddit dominate due to their open APIs and high volume of mental health discourse, other sources like Weibo have been utilized in research focused on Chinese-speaking populations.

Transforming this raw content into structured datasets requires effective annotation strategies, which play a crucial role in determining model performance and generalizability. Table 3 summarizes common annotation approaches identified across the literature.

Table 3. Annotation strategies used in social media mental health research.

| Annotation Strategy | Description / Example | Related Studies |
|---|---|---|
| **Expert Annotation** | Clinicians manually label posts by depression severity or themes (e.g., Beck's Depression Inventory, qualitative coding). | Naseem et al. (2022); Lachmar et al. (2017) |
| **Self-Disclosure Based Labeling** | Uses posts from users who explicitly self-identify as depressed or at-risk to construct datasets (e.g., depressive tweet repositories). | Farruque et al. (2024); Alhamed et al. (2024) |
| **Semi-Supervised Bootstrapping** | Iterative expansion of labeled datasets starting from small expert-labeled seed sets. | Farruque et al. (2024) |
| **LLM-Assisted Annotation** | Zero-shot or few-shot LLMs classify suicidal ideation or depression, extract evidence, or generate synthetic data for augmentation. | Alhamed, Ive, and Specia (2024b); Bucur (2024); Nikmehr et al. (2025); Yang et al. (2023, 2024) |
| **Topic Modeling and Ontology-Guided** | LLM-based discovery of latent mental health themes and annotation guided by PHQ-9 or SNOMED-CT. | Dalal et al. (2025); Zhao and Chen (2025) |

Expert annotation remains a widely adopted method, where clinicians manually label posts according to clinically validated criteria. For example, (Naseem et al. 2022) categorized content by depression severity using Beck's Depression Inventory, while (Lachmar et al. 2017) employed qualitative coding to extract themes from tweets tagged with #MyDepressionLooksLike. Though resource-intensive, this method ensures high-quality labels grounded in psychiatric standards.

In contrast, self-disclosure strategies rely on users who explicitly identify themselves as experiencing mental health conditions. (Farruque et al. 2024) constructed a depressive tweet repository based on such self-identifying users, while (Alhamed, Ive, and Specia 2024a) analyzed language changes before and after self-reported depression diagnoses. These approaches offer scalable alternatives to expert annotation, though they may introduce bias due to varying levels of openness across user groups.
To bridge the gap between scalability and clinical precision, semi-supervised bootstrapping has been applied. These methods start with a small set of expert-labeled data and use model predictions to iteratively expand the training set. This technique enables large dataset construction with minimal manual input.

LLM-assisted annotation is a rapidly emerging approach. For example, (Nikmehr et al. 2025) used zero-shot prompting to detect suicidal ideation without labeled training data, while (Alhamed, Ive, and Specia 2024b; Yang et al. 2023, 2024) applied LLMs to extract explanations and supporting evidence from existing corpora. (Bucur 2024) further demonstrated the utility of synthetic data generation to enhance classification in low-resource scenarios. These models also support higher-level functions such as topic modeling and ontology-guided annotation, as seen in (Zhao and Chen 2025) and (Dalal et al. 2025).

Together, these diverse annotation strategies reflect a growing sophistication in the preparation of social media datasets for mental health research. As LLMs become more capable, hybrid pipelines that combine expert knowledge, user signals, and automated augmentation are expected to define the next phase of dataset construction.

#### *4.1.3 Model Landscape and Evaluation Approaches*

The modeling landscape for social media-based mental health analysis has evolved significantly over the past decade, which reflects advances in natural language processing and a growing interest in deriving mental health insights from online discourse. This section summarizes the dominant modeling paradigms, their roles in mental health classification, and the technical progression observed across reviewed studies.

Early efforts primarily employed traditional machine learning models such as logistic regression, random forest, and LightGBM. These models offered strong baseline performance and clear interpretability but required hand-crafted features and struggled with the informal, context-dependent nature of social media text. As researchers moved toward more flexible architectures, deep learning models like ALBERT and GRU emerged, which could learn features automatically from raw text. These models generally improved classification performance but at the cost of reduced transparency and greater data demands.

A major shift occurred with the adoption of transformer-based models, which enabled more sophisticated linguistic understanding through attention mechanisms and pre-trained contextual embeddings. Studies reviewed in this category introduced architectures such as PSAT, DORIS, Sentiment-guided Transformer, and SpanPHQ. These models expanded the scope of analysis from binary classification to more finegrained tasks, such as estimating emotional severity and detecting symptom-related expressions. Many models also incorporated external domain knowledge during finetuning, enhancing their ability to distinguish between clinically relevant expressions and general discourse. Notably, transformer models enabled a more modular approach to modeling—allowing researchers to tailor model components to specific detection tasks while using pre-trained language backbones.

To address the challenge of limited labeled data, especially in underrepresented populations, many studies employed semi-supervised and bootstrapping strategies. These methods typically began with small, high-quality datasets and expanded them using model-assisted predictions or user-generated signals. For example, depressive tweet repositories and Reddit timelines from self-identified users served as seeds for iterative data augmentation. These techniques improved the scalability of model training while maintaining reasonable quality control through clinician-informed initialization or filtering.

In more recent work, LLM-assisted pipelines have become increasingly prominent. Rather than serving only as classification engines, large language models such as GPT3.5, Meta Llama, and MentaLLaMA are now being used to support annotation, training data expansion, and interpretation. For example, LLMs have been used to generate synthetic posts to enhance low-resource datasets, extract plausible supporting evidence from raw posts, and convert model outputs into natural language summaries. This versatility expands their utility beyond static prediction tasks, and enables broader integration into the model development pipeline.

The diversity of models across studies—ranging from lightweight classifiers to complex transformer ensembles—reflects both the heterogeneity of social media data and the variety of mental health tasks under investigation. These approaches lay the groundwork for more targeted applications, whether for monitoring population-level signals, generating labeled datasets, or prototyping lightweight tools for real-time analysis.

A comprehensive summary of the models, their data sources, and their contributions
to specific tasks is presented in the Table 4 below.

Table 4.: Summary of Models in Social Media Analysis

| Group | Model Type / Name(s) | Data Source(s) | Role / Application | Key Features / Contribution | Related Studies |
|---|---|---|---|---|---|
| **Classical ML and Hybrid Models** | Logistic Regression, Random Forest, LightGBM, ALBERT, GRU | Twitter, Reddit | Depression / Anxiety classification | Compares ML vs DL trade-offs (interpretability vs complexity) | Ding et al. (2025) |
| | Random Forest (dual Twitter model) | Twitter | Early depression detection | Time-aware features from social networks | Cacheda et al. (2019) |
| | Sentiment lexicons + Twitter Gradient Boosting | Twitter | Depression detection | Combines lexicons and content-based features | Chiong et al. (2021) |
| | Bayesian nonparametric hierarchical models | Clinical (LIFE study) | Bipolar disorder course classification | Time-series latent class discovery | Cochran et al. (2016) |
| **Transformer / LLM-Enhanced Detection** | PSAT (knowledge-infused cross-attention with GPT) | Reddit, Twitter | Depression detection with explanations | Uses PHQ-9/SNOMED-CT as external knowledge for explainability | Dalal et al. (2024) |
| | DORIS (LLM + classifiers) | Reddit | Depression detection | LLM annotation, medical criteria, mood course summarization | Lan et al. (2024) |
| | Sentiment-guided Transformer | Reddit, Twitter | Depression severity detection | Combines sentiment with semantic signals using contrastive learning | Zhang et al. (2023) |
| | PHQ-aware SpanPHQ model | Twitter, Reddit | Symptom detection | Combines PHQ-9 guided span-prediction and contrastive learning | Zhang et al. (2023) |
| | LLM-derived embeddings + classifiers | Reddit | Depression detection | Uses LLM-generated summaries as embeddings | Kim et al. (2025) |
| | Fine-tuned GPT-3.5 Turbo and LLaMA2-7B | Twitter | Depression detection | Fine-tuned LLMs for improved accuracy | Shah et al. (2025) |
| | Zero-shot LLMs (GPT, Bard) | Reddit | Post-diagnosis vs pre-diagnosis | Compares user text before/after diagnosis | Alhamed et al. (2024) |
| | LLM-based evidence extraction (Meta Llama) | Reddit | Suicide risk justification | Extracts textual evidence for risk label | Alhamed et al. (2024) |
| | MentaLLaMA (LLM instruction-tuned) | Multi-source | Interpretable mental health analysis | Multi-task explanations with few-shot prompts | Yang et al. (2024) |
| **Semi-supervised and Bootstrapping Approaches** | Semi-supervised LLM-driven SSL | Twitter | Symptom detection | Combines zero-shot with clinician-labeled seeds | Farruque et al. (2024) |
| | GPT-3.5 synthetic data + MPNet | Reddit | Symptom detection (eRisk) | Generates synthetic posts aligned with BDI-II | Bucur et al. (2024) |
| | Zero-shot prompting with LLMs | Twitter, Reddit | Suicidal ideation detection | Detects risk in data-scarce settings without labels | Nikmehr et al. (2025) |
| | LLM-assisted data annotation & augmentation (GPT-4) | Multi-source (social media datasets) | Health-related text classification | Uses zero-shot + augmentation to bootstrap semi-supervised training | Guo et al. (2024) |
| **Multimodal Models** | SenseMood (CNN + BERT multimodal) | Twitter (text + images) | Multimodal depression detection | Combines visual and textual deep features | Lin et al. (2020) |
| | Multimodal dictionary learning | Twitter | Depression detection | Multi-feature fusion: behaviors + clinical features | Shen et al. (2017) |
| **Trend and Qualitative Analyses** | LLM-powered topic modeling | Reddit | Trend analysis | Discovers latent mental health themes | Zhao et al. (2025) |
| | Thematic coding | Twitter | Public discourse themes | Identifies thematic patterns | Lachmar et al. (2017) |
| | Qualitative self-disclosure analysis | Reddit | Behavioral analysis | Impact of social network on self-disclosure | Wang et al. (2025) |

#### *4.1.4 Emerging Trends and Research Directions*

Recent studies in social media-based mental health analysis reflect a shift from broad categorical labeling to more granular, symptom-level modeling. This trend aligns mental health predictions with established diagnostic criteria by identifying specific indicators—such as insomnia or suicidal ideation—rather than general conditions. Such approaches offer increased clinical relevance and enable more precise characterization of user mental states.

Improving model interpretability is also gaining attention. Given the informal and diverse nature of social media language, recent work has explored structured outputs such as annotated symptom spans or text rationales to make predictions more transparent. Models enriched with clinical lexicons and ontologies have proven effective in this regard, especially in enhancing alignment with human review. Instruction-tuned LLMs have emerged as flexible tools in low-resource research settings. Their ability to generalize across tasks without extensive labeled data has facilitated the analysis of emerging linguistic patterns and underrepresented user populations. Studies increasingly apply these models for zero-shot and few-shot classification, particularly where rapid adaptability is needed.

In some cases, LLMs are integrated into hybrid workflows, where they serve as feature generators or preliminary annotators feeding into lighter classifiers. This modular use of generative models reflects an effort to balance scalability with interpretive clarity.

Finally, an increasing number of studies are exploring temporal modeling approaches to track mental health changes over time. These efforts seek to capture language shifts, posting frequency, and thematic evolution as indicators of psychological trajectories. While still in early stages, such longitudinal analyses reflect growing interest in the dynamic nature of online mental health expression.

Together, these trends demonstrate a broader methodological expansion in generative AI applications—moving beyond prediction to support more context-sensitive and clinically attuned mental health research. In the surge of AI technologies in medicine, LLM-powered conversational agents are prime examples of unprecedented tools that offer individualized, highly adaptable, and accessible support for both patients and healthcare providers (Abbasian et al. 2025; G´omez et al. 2024; Ni and Jia 2025).

### *4.2 LLM-Powered Conversational Agents Across Medical Practice Stages*

These highly effective and helpful LLM-Powered agents assist with various stages of medical practice overall, while some provide more specialized support for mental health issues (Cho et al. 2023; Lozoya et al. 2025; Ma, Mei, and Su 2024; Rollwage et al. 2022; Yu and McGuinness 2024). While LLM-powered conversational agents designed for mental health address unique challenges within this specific domain of healthcare, aiming to provide effective and personalized responses to each patient's inquiries while demonstrating remarkable capabilities with challenges present at the same time (Cho et al. 2023; Li et al. 2023; Ma, Mei, and Su 2024). The whole process is illustrated in Figure 2.

LLM-powered conversational agents can improve the triaging of patients (Gaber et al. 2025; Masanneck et al. 2024). Clinical conversational agents powered by LLMs have shown effectiveness in assisting triage decisions. For instance, advice given during conversations by those agents can help reassure users' confidence and accuracy in judgement making during triaging (G´omez et al. 2024). In mental health contexts, (Taylor et al. 2024) assists triage by processing great volumes of unstructured clinical data from electronic health records (EHRs) and analyzing this rich data to offer recommendations for triage; (Thotapalli et al. 2025) uses chatGPT directly to provide feedbacks during youth mental health emergency triage, and found that while the responses given were correct, clear and comprehensive, the slight tendency of over-triaging shows the need for further refinement of the model.

LLM-powered clinical conversational agents have shown to be able to improve traditional symptom checking using its various strengths including emotional support and great adaptability (G´omez et al. 2024; Lawrence et al. 2024; Scholich et al. 2025). In terms of symptom checking, many agents including STEF agent offer personalized, empathetic responses geared towards users' conditions (Jin et al. 2025). While many models are powerful enough in checking symptoms and diagnosis, (Lawrence et al. 2024) mentions that accuracies can be further improved by specifying the accurate diagnostic category. In addition, some models are trained in more specific domains to achieve higher symptom checking accuracy, such as (Ji et al. 2022) for detecting depression and suicidal ideation.

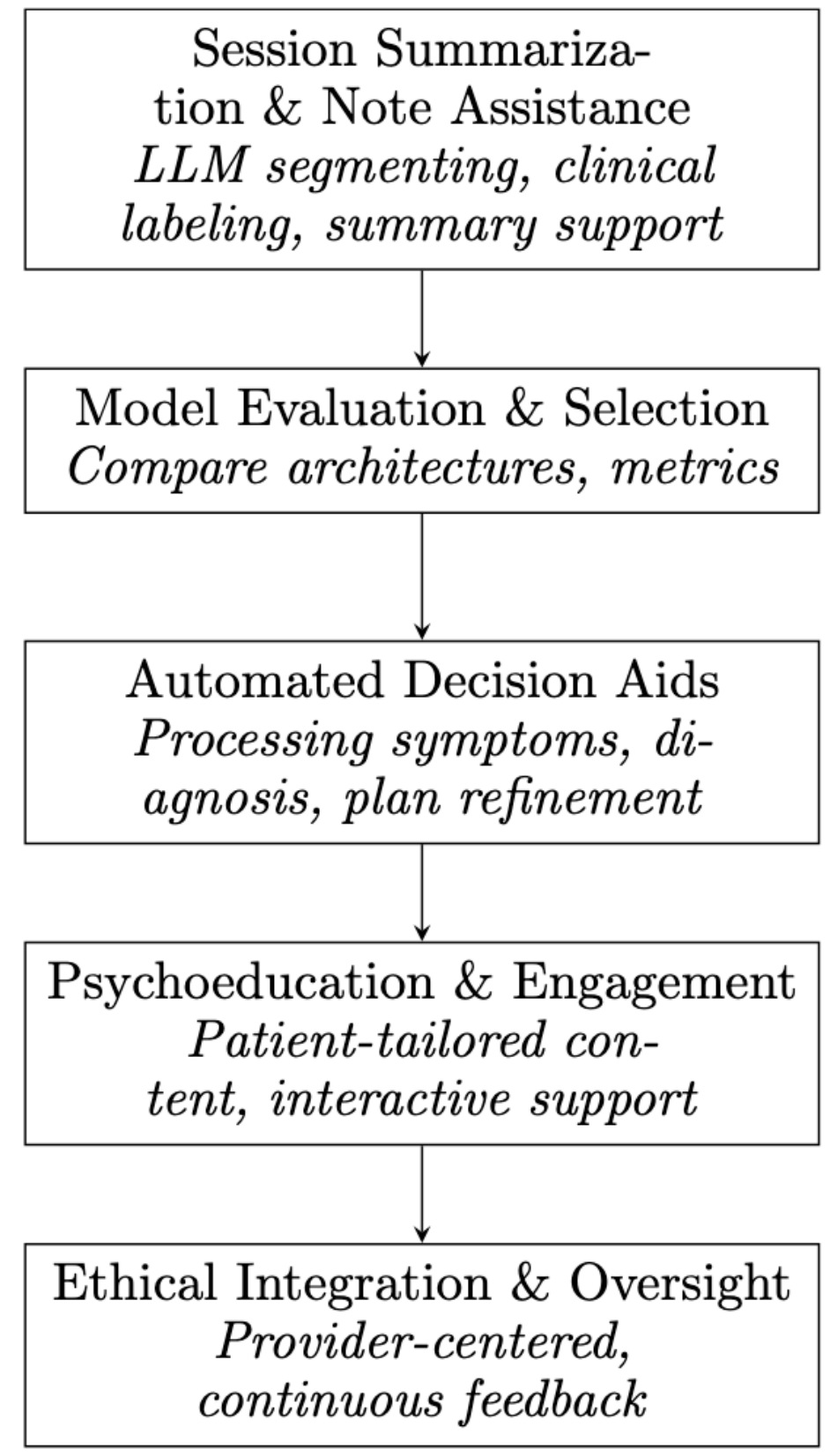

Figure 2. Sequential roles of LLMs in therapy support: from session summarization and model evaluation through to decision aids, patient engagement, and ethical oversight.

Figure 2 is a flow chart illustrating the sequential roles of LLMs in supporting therapy, from session summarization and model evaluation to aiding clinical decisions, enhancing patient engagement, and ensuring ethical oversight. A detailed explanation of each component follows. Patient engagement has emerged as one of the most significant benefits of LLM-powered conversational agents. (Cevasco et al. 2024; Ma, Mei, and Su 2024; Wen et al. 2024). Conversational agents like ChatGPT have been shown to greatly improve patient engagement though bonds building that are close to real human interactions, which results in patients' increased willingness to engage with conversational agents (Cevasco et al. 2024). When it comes to mental health problems, the non-judgmental, immediately accessible support from these conversational agents is crucial for various users with issues ranging from stigma to limited access to therapists (Ma, Mei, and Su 2024). Moreover, (Wah 2025) shows that integrating LLM capabilities with human insights can further promote patient engagement compared to AI-only models.

LLMs are increasingly recognized as valuable tools in mental health therapy, offering a wide range of support that spans multiple stages of therapy sessions.

### *4.2.1 Session Summarization and Clinical Note Assistance*

LLMs have proved significant capabilities in summarizing therapy sessions by automating the process from multiple perspective. (So et al. 2024) demonstrates that fine-tuned LLMs effectively delineate symptom-related dialogue segments and label them with clinical categories. Moreoever, the models produce coherent, clinically relevant summaries that describe symptoms. Those summaries help health professionals with supporting their patient more efficiently using content highlighted by LLMs.
In terms of the technical aspects of models being used, (Sahu et al. 2025) fine-tunes and evaluates five well-known pretrained summarization models (BART-base (Lewis et al. 2020), BART-large-CNN (Lewis et al. 2020) which is BART-base fine-tuned using CNN Daily mail dataset (Hermann et al. 2015), T5-large (Raffel et al. 2020), BART-large-xsum-samsum (Gliwa et al., 2019), Pegasus-large (Zhang et al. 2020)) using dataset collected by them covering diverse mental health aspects. Pegasus-large (Zhang et al. 2020) has scores highest in several metrics indicating better summaries while BART-large-CNN has better generalizability (Sahu et al. 2025). In addition, some studies found that decoder-only LLM architectures have advantages over encoder-decoder models in their summarization tasks (Adhikary et al. 2024). Meanwhile, while task-specific models fine-tuned on mental health data (Ji et al. 2022; Yang et al. 2024) perform well more as expected, some LLM that were not specifically trained in mental health data, performed comparably and was rated highest across six parameters used (Adhikary et al. 2024).

It is worth mentioning that ethical concerns are highlighted in multiple paper as these LLMS are tools to assist mental health professionals instead of substitutes to professional healthcare providers, especially in low-resource settings (Sahu et al. 2025; So et al. 2024). More details about ethics are discussed in Section 5.2.

### *4.2.2 Decision Aids for Treatment Planning*

Due to the capability of LLM to interpreting various formats of material including text and audio, LLMs can be applied to automatically extract and process information symptoms in clinical discussion (Bouguettaya, Stuart, and Aboujaoude 2025) as discussed in above section. Using information from clinical discussion recorded and significant medical databases, LLMs also helps with further refining treatment (Bouguettaya, Stuart, and Aboujaoude 2025). The overall clinical workflow integrating triage, symptom checking, patient engagement, session summarization, and treatment planning is illustrated in Figure 3.

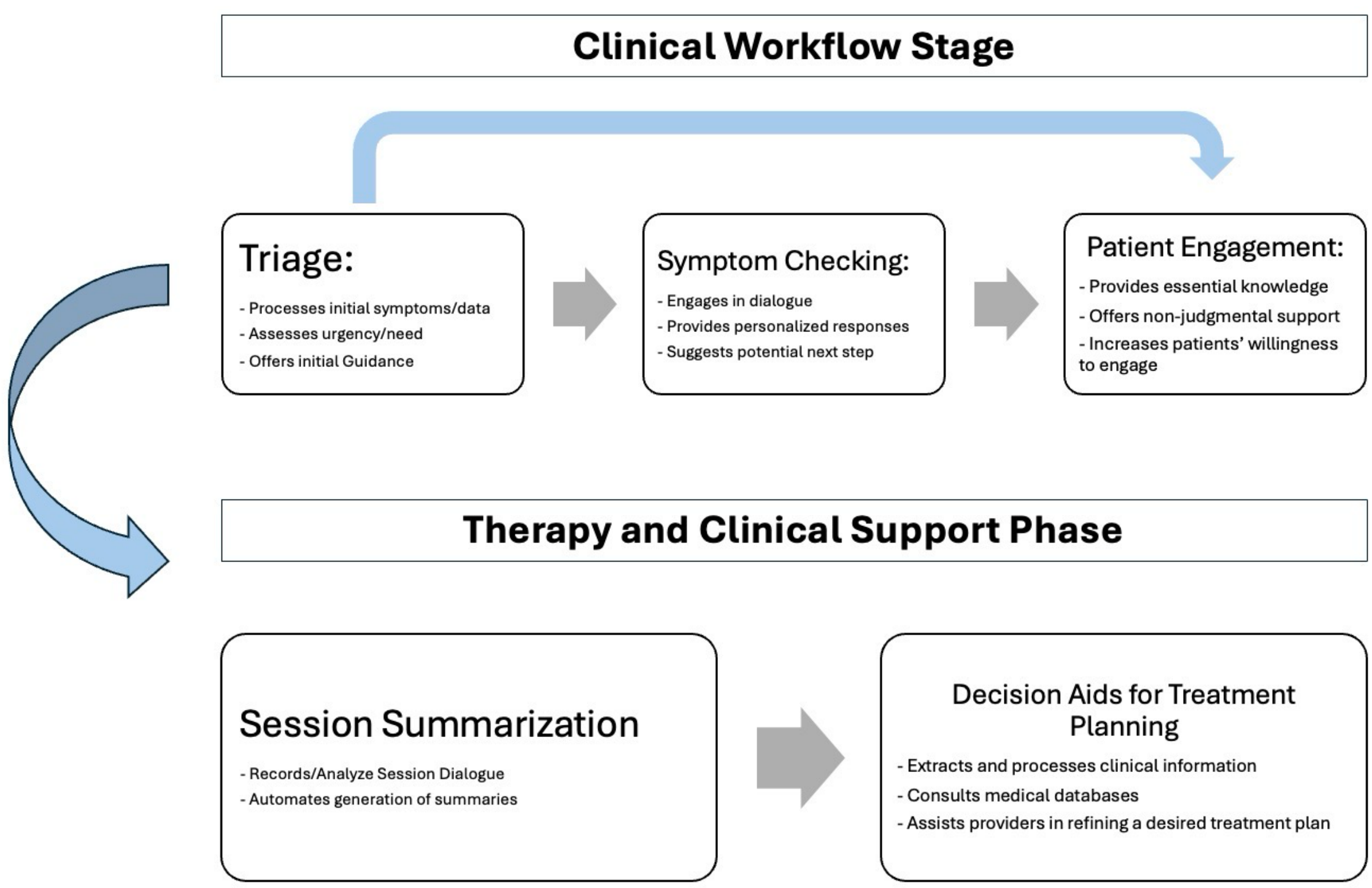

Figure 3. Clinical workflow and support phases integrating triage, symptom checking, patient engagement, session summarization, and decision aids for treatment planning in mental health care.

In the context of mental health, (Obradovich et al. 2024) further specifies the prospects and potentials of LLMs in mental health care. To illustrate, LLMs has shown potential in understanding verbal information to deduce hidden causes and influences, introducing a new perspective of evaluating patients' mental health conditions (Obradovich et al. 2024). Additionally, these benefits of LLMs are not restricted to fine-tuned models but also present in LLMs trained on general corpora that includes significant amount of medical language but not specific to medicine (Obradovich et al. 2024). In order to achieve optimal results, (Bouguettaya, Stuart, and Aboujaoude 2025) points out that ideally a provider should be allowed to input patient's information to LLMs to receive a timely report that encapsulate related diagnosis to help with developing a desired treatment plan.

By offering customized, accessible information, LLMs can help healthcare professionals efficiently with their decision-making process while not taking over their important judgment and show great potential in assisting treatment planning (Jin et al. 2025; Lawson McLean et al. 2024).

*4.2.3 Psychoeducational Content and Patient Engagement*

As a fundamental aspect of therapy, psychoeducation aims to equip patients and their families with essential knowledge about mental disorders, treatment options, coping mechanisms, and prevention strategies. In general, digital mental health interventions (DMHIs) have been explored as an approach of psychoeducation. Even though DMHIs have strong prospects in improving accessibility of psychoeducation, but challenges remain in patient engagement and maintaining high percentages of patients completing treatment (Cross and Alvarez-Jimenez 2024)

In comparison to more conventional DMHIs, LLM-powered conversational agents have shown to improve patient engagement through generating understanding and interactive conversations. (Wang et al. 2025)

### *4.3 LLM Prompt Engineering Techniques for Mental Health Applications*

Prompt Engineering is a relatively new discipline for directly use LLMs for a wide variety of applications across diverse domains. Using prompt engineering skills can help better understand the capabilities of LLMs. Prompt Engineering generally refers to the systematic design, formulation and adjustment of input texts to inspire LLM to generate accurate and relevant responses. In the mental health domain, prompt engineering is emerging as a vital technique for tailoring general-purpose LLMs to the sensitive and high stakes requirements of psychological assessment and disease intervention. Figure 4 summarizes our overview for Prompt Engineering Techniques for Mental Health Applications.

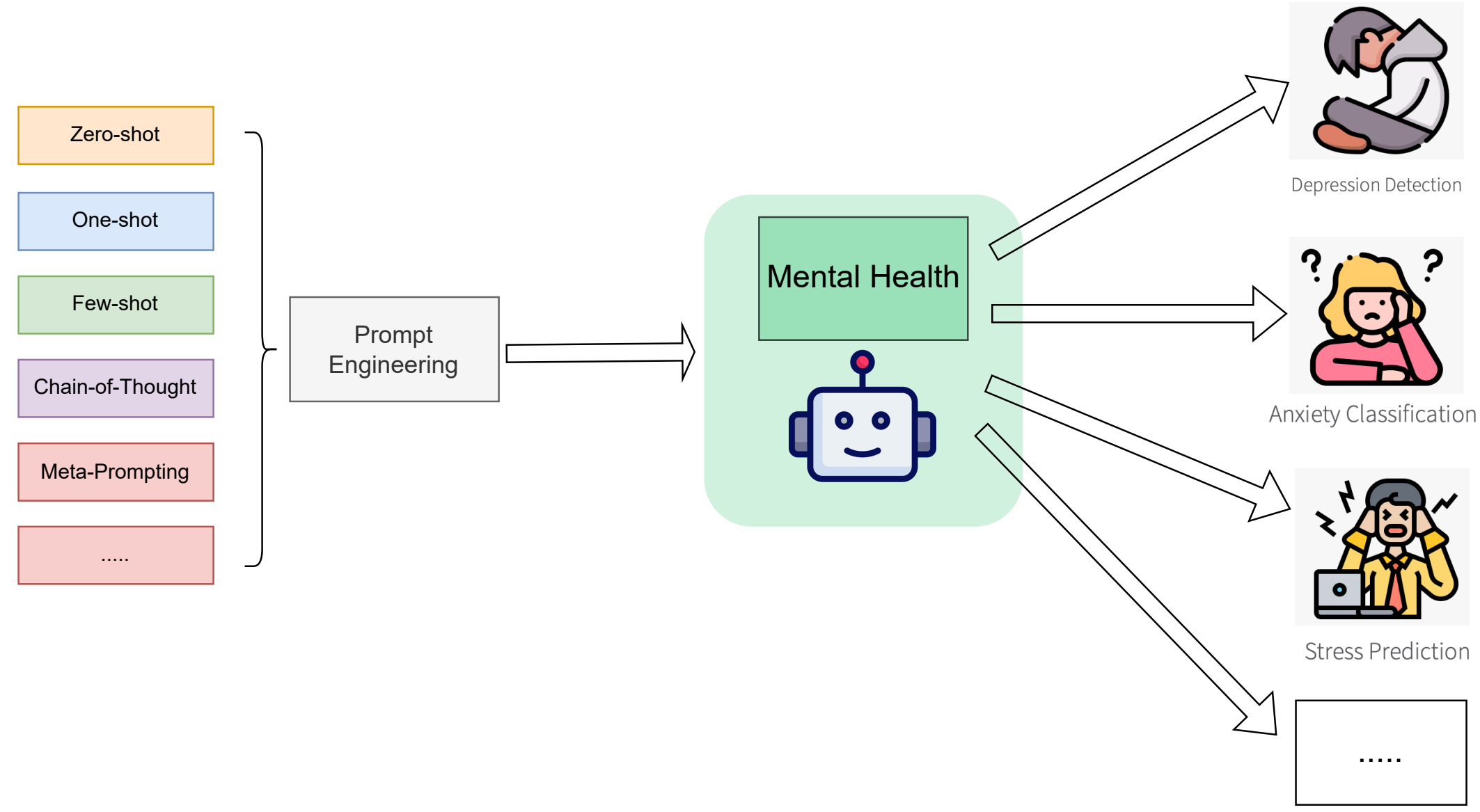

Figure 4. Prompt Engineering Techniques for Mental Health

#### *4.3.1 Overview of Prompt Engineering in Mental Health*

Unlike extensive LLM fine-tuning, which requires large-labeled datasets and computational resources, prompt engineering offers a lightweight and flexible pathway to effectively instruct LLMs for domain-specific mental health tasks. Well-designed prompt template can guide LLMs to perform structured screening such as depressive symptom extraction (Teferra et al. 2025). Also, to simulate psychotherapy and psychoeducational content delivery, there are also some previous works to explore how to guide LLM to improve related heath tasks (Lozoya et al. 2025), which proposed a Client101 platform to generate personalized health advice according to user psychotherapy medical history. Therefore, here we aim to explore and summarize the overview of prompt engineering in mental health.

### *4.3.2 Methods and Paradigms of Prompt Engineering*

Prompt engineering has rapidly become a cornerstone in harnessing the potential of LLMs for health and medical tasks, enabling highly efficient adaptation of foundation models to domain-specific needs without costly and time-consuming retraining (Mesk´o 2023). In digital mental health, prompt engineering supports a range of tasks, from automated symptom classification to simulated therapeutic dialogue and question-answer chatbots. Recent reviews highlight that these methods are particularly effective for early screening, information extraction, and patient engagement in resource-constrained settings (Priyadarshana et al. 2024; Wang et al. 2023; Zaghir et al. 2024).

Classical prompt engineering methods in health AI include zero-shot, one-shot, and few-shot prompting. In zero-shot prompting, the LLMs only receive the task description and will generate the answer using the pretrained knowledge including general corpus (Zaghir et al. 2024). For few-shot and one-shot learning give some curated examples about the question and enable LLMs to ground its reasoning (Oniani et al. 2023). More advanced paradigms such as Chain-of-Thought (CoT) prompting require the model not just to provide an answer, but to explicitly reason step by step. CoT methods with self-consistency, which aims to instruct LLMs to try multiple reasoning chains and aggregate outcomes, have shown promising potential in complex mental health diagnostic questions and context-sensitive counseling simulations (Wang et al. 2022; Xu et al. 2024b). Meta-prompting which means using initial prompt to generate new prompts, prompt chaining for organizing multi-step task-specific prompts and generated knowledge prompting (enriching prompts with information synthesized by the LLM itself or external tools) are also some excellent methods to repurpose LLMs to solve complex questions (Chan et al. 2025; Zaghir et al. 2024).

### *4.3.3 Application Scenarios in Mental Health*

In mental health, there are always some application scenarios that need to analyze complex medical health notes and extract the key factors that can identify the targeted mental health issue. For example, LLMs are effectively utilized for early detection and classification of mental health conditions, such as depression diagnosis and flourishing classification (Kim et al. 2024; Lawrence et al. 2024). Specifically, by analyzing unstructured data sources like social media posts and electronic medical records (EMRs), LLM can identify linguistic and behavioral indicators tied to mental health states. There are some studies that point out models like GPT-4 and domain-adapted variants such as MentaLLaMA and PsychBERT can achieve competitive accuracy compared to traditional approaches in symptom recognition and risk stratification (Vajre et al. 2021; Wang et al. 2023). In clinical treatments and interventions, LLMs can assist health experts by extracting key symptom information from much clinical visits records and enhance clinical decision-making processes. They contribute to personalized treatment planning and provide cognitive behavioral therapy aids, such as reframing negative thoughts (Ferrario, Sedlakova, and Trachsel 2024; Obradovich et al. 2024). Also, as for mental health counseling and education, LLMs also play an important role. In detail, LLMs generate psychoeducational materials and support mental health resource supplementation. Such automated content tools show promise in bridging gaps in mental health literacy and reducing gaps related to stigma and access.

In short, there are much advanced developments for LLM applications in mental health. With ongoing challenges, we can still explore how to effectively adapt LLMs for mental health.

## 5. Innovations and Challenges

Recent advances in multimodal learning and LLMs are reshaping the landscape of digital mental health. Multimodal frameworks integrate heterogeneous data such as text, speech, physiological signals, and behavioral patterns, which will offer more comprehensive representations of psychological states than single modality approaches. At the same time, LLM-driven systems extend these advances by enabling scalable interaction, monitoring, and moderation in online platforms. However, these innovations are coupled with significant ethical and sociotechnical challenges, including risks of misinformation, privacy violations, and accountability in AI-mediated care. This section examines both the methodological innovations that expand analytic and clinical capacities, and the challenges that must be addressed to ensure the responsible deployment of these technologies in mental health contexts.

### *5.1 Multimodal Learning for Mental Health*

Multimodal learning is a promising direction for processing heterogeneous data sources such as text data, speech data and time series data as shown in Figure 5. Compared to traditional methods relying on single data modality, multimodal frameworks enable integrating complex information simultaneously and provide information loss within diverse modality data (Li and Tang 2024; Li et al. 2020; Stahlschmidt, Ulfenborg, and Synnergren 2022).

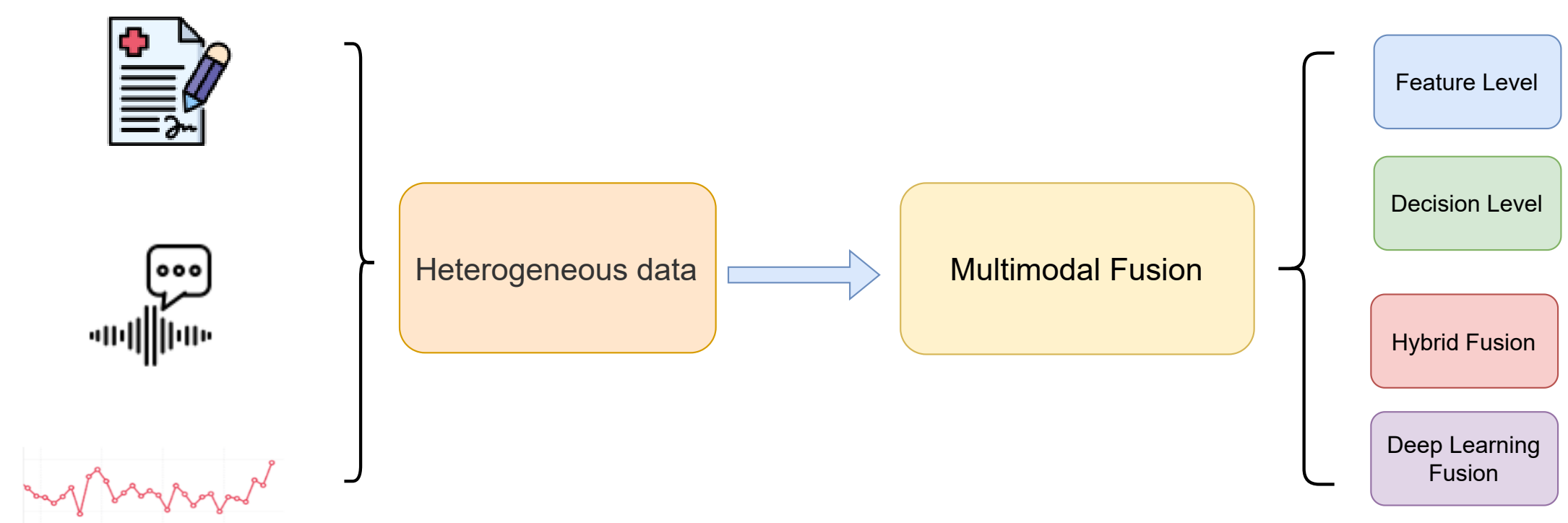

Figure 5. Multimodal Learning for Mental Health

The complexity of mental health is reflected in its multi-dimensional and multi-layered manifestations, which a single modality often cannot fully capture to represent an individual's true psychological state. In recent years, there are many studies on multimodal data fusion that has rapidly advanced in mental health application. Generally, multimodal data mainly include patients' linguistic texts such as clinical interview screening reports, voice data such as speech intonation and pauses, and daily behavioral sensor data including step count and conversation duration (Jin, Ye, and Li 2025; Sahili, Patras, and Purver 2024; Sharma et al. 2025). These heterogeneous data complement each other to comprehensively reflect multiple aspects of mental states. Therefore, effective fusion of different modality data is important for mental health tasks, such as depression diagnosis and flourishing classification (Pillai et al. 2025). Table 5 summarizes the representative modal modality in mental health.

Table 5. Mental Health Multimodal Data.

| Modality | Main Characteristics | Mental Health Applications | Related Studies |
|---|---|---|---|
| **Text** | Clinical interviews, social media posts, therapy notes; capture lexical, syntactic, and semantic cues. | Depression screening, suicidal ideation detection, therapy monitoring. | Yazdavar et al. (2020); Sahili et al. (2024); Scherbakov et al. (2025) |
| **Audio** | Speech recordings and phone calls; includes prosody, speech rate, and voice quality features. | Depression and bipolar disorder detection, stress and emotion analysis. | Poria et al. (2017); Sadeghi et al. (2024) |
| **Video** | Facial expressions, eye gaze, head pose, and body movements. | Emotion recognition, PTSD detection, psychomotor retardation assessment. | Jiang et al. (2023); Jin et al. (2025) |
| **Physiological** | Data from wearables (heart rate variability, skin conductance), EEG, or fMRI. | Stress quantification, anxiety assessment, and sleep quality monitoring. | Jiang et al. (2024); Song et al. (2024) |
| **Behavioral** | Activity tracking and social interaction frequency via smartphones or wearables. | Long-term mood monitoring, relapse prediction, and ecological momentary assessment. | Chen et al. (2024); Yazdavar et al. (2020) |

As illustrated in Table 6, multimodal fusion techniques integrate diverse data sources to address complex mental health tasks and result in more accurate diagnosis, symptom prediction, and personalized interventions. Fusion methods are typically categorized according to the stage at which multimodal data are combined—such as feature-level, decision-level, hybrid, or deep learning–based fusion strategies.

Table 6. Fusion Paradigms in Multimodal Mental Health Modeling

| Fusion Paradigm | Description | Suitable Task Types | Related Studies |
|---|---|---|---|
| **Feature-Level Fusion** | Concatenates or integrates features from different modalities into a single feature vector before modeling. | Tasks needing joint feature representation, e.g., emotion recognition. | Chen et al. (2015); Zhang et al. (2025); Cheng et al. (2024) |
| **Decision-Level Fusion** | Each modality is modeled separately; predictions are combined via voting, averaging, or stacking. | Classification tasks where modality predictions can be combined easily. | Almeida et al. (2024); Calhoun et al. (2016) |
| **Hybrid Fusion** | Combines intermediate representations using attention mechanisms or shared latent spaces to enable cross-modal interaction. | Suitable for recommender systems and filtering tasks. | Shalu et al. (2020); Nayak et al. (2025) |
| **Deep Learning-Based Fusion** | Uses specialized neural architectures (e.g., transformers, graph neural networks) to dynamically model spatial– | Temporal pattern analysis, sequential behavior prediction, multi-scale feature | Rahaman et al. (2022); Sahili et al. (2024); |

| | temporal and inter-modal relationships. | integration, and dynamic relationship modeling. | Zhang et al. (2024) |
|---|---|---|---|

### *5.2 Ethical, and Sociotechnical Considerations*

The integration of LLMs into mental health-related social media platforms introduces a complex array of ethical and sociotechnical risks that extend beyond model accuracy or performance. One pressing concern is the use of LLMs in social media influence operations, where these models can generate emotionally persuasive and seemingly authentic messages designed to manipulate user beliefs. Such tactics pose unique dangers in mental health contexts, where individuals may be more vulnerable to suggestive or emotionally charged content. These influence operations often exploit the naturalistic fluency of LLM-generated text, which makes them difficult to detect or attribute, and enables covert dissemination of stigmatizing narratives or harmful ideologies under the guise of support or community interaction (Dalal et al. 2025). Further complicating this landscape is the challenge of detecting when content has been generated by an LLM at all. As recent work has shown, users and moderators alike can become "lost in transformation," unable to distinguish real peer support from synthetic campaigns—undermining trust in digital mental health spaces (Grimme et al. 2023). The ease with which LLMs can scale and adapt language for particular psychological triggers has heightened the risk of misinformation propagation, especially when fine-tuned on biased or manipulative data sources. This has prompted an emerging body of work aimed at developing mitigation strategies, including content auditing, watermarking, and model behavior analysis to counteract the misuse of LLMs in vulnerable populations (Zhang et al. 2024a).

In parallel with these risks, the use of LLM agents as autonomous moderators in decentralized or lightly governed platforms introduces further tensions between efficiency and accountability. For example, LLMs have been proposed as tools for enforcing community rules and safeguarding user interactions in decentralized mental health forums. While these agents can reduce the burden on human moderators, they may also enforce rules inconsistently or misclassify sensitive expressions as violations, particularly when mental health disclosures do not conform to normative linguistic patterns (La Cava and Tagarelli 2025). Moreover, the implicit authority granted to such agents—especially when they intervene in emotionally fraught conversations—raises questions about their ethical alignment with community values and the users' right to contest automated decisions.

Privacy and self-disclosure add another layer of ethical complexity. Mental health communities often foster environments that encourage users to share personal struggles, emotions, and behavioral patterns. LLMs trained to interpret and respond to such disclosures may inadvertently compromise user privacy, especially when embedded in models that perform social network inference or behavioral profiling. A recent study demonstrated that LLMs could extract and recontextualize self-disclosed content from online depression communities (Wang and Lu 2025). These concerns are echoed in applied settings such as eating disorder recovery, where LLM chatbots act simultaneously as therapeutic companions and potential triggers. In these cases, the very same empathetic interface that enables support can also normalize harmful behaviors or reinforce negative self-concepts. This dual role reveals a critical dilemma: the systems designed to promote recovery can, without proper constraints, also exacerbate risk (Choi et al. 2025).

To address these challenges, recent scholarship has proposed frameworks that articulate the specific ethical dilemmas posed by LLM-driven mental health applications. One such review identifies key concerns including the illusion of understanding, false empathy, and the lack of accountability when AI-generated advice goes wrong. It urges a reevaluation of how therapeutic responsibility is assigned when LLMs enter the domain of care work, especially in situations where users may place trust in systems that are not explicitly therapeutic or clinician-supervised (Cabrera et al. 2023). The SouLLMate framework offers a structured design for adaptive LLM deployment that emphasizes contextual awareness, clinical alignment, and multi-level oversight. It highlights the need for mental health systems to move beyond static risk checklists and instead embed ethical reflexivity into the development and deployment cycles of LLM agents (Guo et al. 2024a).

Taken together, these insights underscore the importance of treating LLMs not merely as technical systems but as sociotechnical actors embedded within broader clinical, cultural, and regulatory environments. As mental health support increasingly shifts into online and hybrid modalities, the ethical deployment of LLMs will require coordinated attention to misinformation safeguards, community governance, disclosure dynamics, and the long-term psychological implications of AI-mediated care. Building truly responsible systems will depend not only on model optimization but also on interdisciplinary collaboration, patient-centered design, and sustained investment in digital mental health ethics.

## 6. Discussion

Our work shows that LLMs are reshaping mental health research and practice through a diverse set of application domains. Social media analysis mainly includes depression and suicidal ideation. By bridging social media analysis with clinical conversational agents and therapy support tools, LLMs enable support that spans from population-level monitoring to individualized care. Prompt engineering presents the techniques to guide LLMs in diverse health applications. Multimodal Learning can achieve fusion across diverse modality data. These advances highlight how LLMs outperform traditional models for screening and intervention, and emphasize their flexibility in tasks like automated record summarization and personalized therapy support. Also, we reflect on cross-domain and multimodal integration efforts, referencing methods that fuse text, audio, and sensor data for more robust assessments and discuss advances in prompt engineering, including recent examples where zero-shot and low-shot LLMs have enabled rapid scalability with minimal labeled data.

However, there are still challenges and gaps as follows. First, most studies continue to rely on small, imbalanced, and convenience datasets, particularly those drawn from social media platforms, which can introduce biases. Such datasets often underrepresent certain demographics or clinical features, which limits generalizability and may perpetuate disparities in mental health care. Data scarcity and poor annotation quality further hinder robust evaluation. Second, model reliability and interpretability remain unresolved challenges. Although attention mechanisms improve transparency, LLMs can still generate inconsistent or even harmful outputs. Also, LLMs exhibit many hallucinations when applying in health tasks, which is due to implicit knowledge and analysis rules that LLMs cannot capture. Therefore, while LLMs demonstrate transformative potential in mental health, but their safe and effective use requires better data and stronger safeguards.

Also, expand on ethical and regulatory considerations: fairness, transparency, potential algorithmic bias, and the critical need for robust accountability frameworks before widely clinical deployment. We need to rethink the importance of maintaining human oversight, as LLMs are not a replacement for clinicians but a complementary tool. Note regulatory challenges regarding direct-to-consumer applications and data privacy.

For future work, progress in mental health technology will depend on building richer and more diverse datasets that truly represent the complexities of different populations and real-world settings. Developing more powerful model can enhance mental health care, enabling clearer insights into how decisions are made. Besides, tools must be resilient and adaptable as environments evolve, with ongoing monitoring to quickly catch and address any issues.

## 7. Conclusion

Large language models (LLMs) have emerged as powerful tools with the potential to transform multiple domains of mental health, from early detection of depression and suicidal ideation on social media to supporting clinical conversation, therapy, and psychoeducation in both digital and traditional care settings. By using multimodal signals ranging from social media posts and electronic medical records to speech and sensor data, LLMs have enabled more nuanced, personalized, and scalable interventions. Advances in prompt engineering and domain adaptation further expand LLM flexibility, which allows rapid deployment and adaptation without the need for extensive retraining or costly data collection. However, the integration of these technologies must be grounded in ongoing dialogue, careful empirical validation, and alignment with the needs of both clinicians and patients, to realize their full promise in improving access, equity, and outcomes in mental health care.

Despite these advances, the field faces enduring methodological, technical, and social challenges that must be addressed to support safe and effective deployment of LLMs in real-world clinical contexts. Data limitations persist, with many studies relying on small, imbalanced, or convenience samples that introduce bias and limit generalizability, especially across diverse populations and underrepresented subgroups. Model reliability, explainability, and risk of hallucinations remain open concerns. All of these challenges urge the use of transparent, clinically validated evaluation frameworks and the development of safeguards to mitigate unintended consequences. Furthermore, integrating LLMs into dynamic mental health environments calls for robust adaptation strategies, longitudinal monitoring, and continuous feedback loops that respond to evolving social and clinical contexts. Only through dedicated efforts to build inclusive datasets and refine task-specific models can the potential of LLMs be translated into tools that reliably augment clinical decision-making and patient care.

Ethical and regulatory challenges are especially significant as language model-based systems become more involved in confidential and critical mental health interactions. To address these issues, it is essential to prioritize fairness, privacy, and transparency, alongside strong accountability protocols, before these technologies see broad usage in clinical environments. Rather than replacing human professionals, LLMs should function as supportive aids with ongoing expert oversight. Continued collaboration across disciplines, community participation, and investment in ethical development are crucial for creating trustworthy AI mental health solutions that benefit society.

## Reference

Abbasian, Mahyar, Iman Azimi, Amir M Rahmani, and Ramesh Jain. 2025. "Conversational health agents: a personalized large language model-powered agent framework." JAMIA Open 8 (4): ooaf067.

Abdmeziem, Mohammed Riyadh, and Amina Ahmed Nacer. 2025. "Leveraging IoT and LLM for depression and anxiety disorders: a privacy preserving perspective." Security and Privacy 8 (4): e70061.

Adhikary, Prottay Kumar, Aseem Srivastava, Shivani Kumar, Salam Michael Singh, Puneet Manuja, Jini K Gopinath, Vijay Krishnan, Swati Kedia Gupta, Koushik Sinha Deb, and Tanmoy Chakraborty. 2024. "Exploring the efficacy of large language models in summarizing mental health counseling sessions: benchmark study." JMIR Mental Health 11: e57306.

Alhamed, Falwah, Julia Ive, and Lucia Specia. 2024a. "Classifying Social Media Users before and after Depression Diagnosis via Their Language Usage: A Dataset and Study." In Proceedings of the 2024 Joint International Conference on Computational Linguistics, Language Resources and Evaluation (LREC-COLING 2024), edited by Nicoletta Calzolari, Min-Yen Kan, Veronique Hoste, Alessandro Lenci, Sakriani Sakti, and Nianwen Xue, Torino, Italia, May, 3250–3260. ELRA and ICCL. https://aclanthology.org/2024.lrec-main.289/.

Alhamed, Falwah, Julia Ive, and Lucia Specia. 2024b. "Using Large Language Models (LLMs) to Extract Evidence from Pre-Annotated Social Media Data." In Proceedings of the 9th Workshop on Computational Linguistics and Clinical Psychology (CLPsych 2024), edited by Andrew Yates, Bart Desmet, Emily Prud'hommeaux, Ayah Zirikly, Steven Bedrick, Sean MacAvaney, Kfir Bar, Molly Ireland, and Yaakov Ophir, St. Julians, Malta, Mar., 232–237. Association for Computational Linguistics. https://aclanthology.org/2024.clpsych-1.22/.

Almeida, Filipe Fontinele de, Kelson Rˆomulo Teixeira Aires, Andr´e Castelo Branco Soares, Laurindo De Sousa Britto Neto, and Rodrigo De Melo Souza Veras. 2024. "Multimodal Fusion for Depression Detection Assisted by Stacking Deep Neural Networks." TechRxiv, http://dx.doi.org/10.36227/techrxiv.171018258.87887016/v1.

Belcastro, Loris, Riccardo Cantini, Fabrizio Marozzo, Domenico Talia, and Paolo Trunfio. 2025. "Detecting mental disorder on social media: a ChatGPT-augmented explainable approach." Online Social Networks and Media 48: 100321.

Berrezueta-Guzman, Santiago, Mohanad Kandil, Mar´ıa-Luisa Mart´ın-Ruiz, Iv´an Pau de la Cruz, and Stephan Krusche. 2024a. "Future of ADHD care: evaluating the efficacy of ChatGPT in therapy enhancement." In Healthcare, Vol. 12, 683. MDPI.

Berrezueta-Guzman, Santiago, Mohanad Kandil, Mar´ıa-Luisa Mart´ın-Ruiz, Iv´an Pau de la Cruz, and Stephan Krusche. 2024b. "Exploring the Efficacy of Robotic Assistants with ChatGPT and Claude in Enhancing ADHD Therapy: Innovating Treatment Paradigms." In 2024 International Conference on Intelligent Environments (IE), 25–32.

Bouguettaya, Ayoub, Elizabeth M Stuart, and Elias Aboujaoude. 2025. "Racial bias in AI-mediated psychiatric diagnosis and treatment: a qualitative comparison of four large language models." npj Digital Medicine 8 (1): 332.

Bril-Barniv, Shani, Galia S Moran, Adi Naaman, David Roe, and Orit Karnieli-Miller. 2017. "A qualitative study examining experiences and dilemmas in concealment and disclosure of people living with serious mental illness." Qualitative Health Research 27 (4): 573–583.

Bucur, Ana-Maria. 2024. "Leveraging LLM-generated data for detecting depression symptoms on social media." In International Conference of the Cross-Language Evaluation Forum for European Languages, 193–204. Springer.

Cabrera, Johana, M Soledad Loyola, Irene Maga˜na, and Rodrigo Rojas. 2023. "Ethical dilemmas, mental health, artificial intelligence, and llm-based chatbots." In International WorkConference on Bioinformatics and Biomedical Engineering, 313–326. Springer.

Cacheda, Fidel, Diego Fernandez, Francisco J Novoa, and Victor Carneiro. 2019. "Early detection of depression: social network analysis and random forest techniques." Journal of medical Internet research 21 (6): e12554.

Cai, Jinyu, Jialong Li, Mingyue Zhang, Munan Li, Chen-Shu Wang, and Kenji Tei. 2024. "Language evolution for evading social media regulation via llm-based multi-agent simulation." In 2024 IEEE Congress on Evolutionary Computation (CEC), 1–10. IEEE.

Calhoun, Vince D., and Jing Sui. 2016. "Multimodal Fusion of Brain Imaging Data: A Key to Finding the Missing Link(s) in Complex Mental Illness." Biological Psychiatry: Cognitive Neuroscience and Neuroimaging 1 (3): 230–244. http://dx.doi.org/10.1016/j.bpsc.2015.12.005.

Cardamone, Nicholas C, Mark Olfson, Timothy Schmutte, Lyle Ungar, Tony Liu, Sara W Cullen, Nathaniel J Williams, and Steven C Marcus. 2025. "Classifying unstructured text in electronic health records for mental health prediction models: large language model evaluation study." JMIR Medical Informatics 13 (1): e65454.

Cevasco, Kevin E, Rachel E Morrison Brown, Rediet Woldeselassie, and Seth Kaplan. 2024. "Patient Engagement with Conversational Agents in Health Applications 2016–2022: A Systematic Review and Meta-Analysis." Journal of medical systems 48 (1): 40.

Chan, Callum, Sunveer Khunkhun, Diana Inkpen, and Juan Antonio Lossio-Ventura. 2025. "Prompt Engineering for Capturing Dynamic Mental Health Self States from Social Media Posts." In Proceedings of the 10th Workshop on Computational Linguistics and Clinical Psychology (CLPsych 2025), 256–267.

Chen, Jing, Bin Hu, Lixin Xu, Philip Moore, and Yun Su. 2015. "Feature-level fusion of multimodal physiological signals for emotion recognition." In 2015 IEEE International Conference on Bioinformatics and Biomedicine (BIBM), 395–399. IEEE.

Chen, Xieling, Haoran Xie, Xiaohui Tao, Fu Lee Wang, Mingming Leng, and Baiying Lei. 2024. "Artificial intelligence and multimodal data fusion for smart healthcare: topic modeling and bibliometrics." Artificial Intelligence Review 57 (4). http://dx.doi.org/10.1007/s10462-024-10712-7.

Cheng, Zixuan, Xisheng Huang, and Yang Ding. 2024. "An intelligent depression detection model based on multimodal fusion technology." Journal of Mechanics in Medicine and Biology 24 (8). http://dx.doi.org/10.1142/S0219519424400463.

Chiong, Raymond, Gregorious Satia Budhi, and Sandeep Dhakal. 2021. "Combining sentiment lexicons and content-based features for depression detection." IEEE Intelligent Systems 36 (6): 99–105.

Cho, Young-Min, Sunny Rai, Lyle Ungar, Jo˜ao Sedoc, and Sharath Chandra Guntuku. 2023. "An integrative survey on mental health conversational agents to bridge computer science and medical perspectives." In Proceedings of the Conference on Empirical Methods in Natural Language Processing. Conference on Empirical Methods in Natural Language Processing, Vol. 2023, 11346.

Choi, Ryuhaerang, Taehan Kim, Subin Park, Jennifer G. Kim, and Sung-Ju Lee. 2025. "Private Yet Social: How LLM Chatbots Support and Challenge Eating Disorder Recovery." In Proceedings of the 2025 CHI Conference on Human Factors in Computing Systems, CHI '25, New York, NY, USA. Association for Computing Machinery. https://doi.org/10.1145/3706598.3713485.

Cochran, AL, MG McInnis, and DB Forger. 2016. "Data-driven classification of bipolar I disorder from longitudinal course of mood." Translational psychiatry 6 (10): e912–e912. Cross, Shane P, and Mario Alvarez-Jimenez. 2024. "The digital cumulative complexity model: a framework for improving engagement in digital mental health interventions." Frontiers in Psychiatry 15: 1382726.

Dalal, Sumit, Sarika Jain, and Mayank Dave. 2024. "Review of advancements in depression detection using social media data." IEEE Transactions on Computational Social Systems. Dalal, Sumit, Deepa Tilwani, Manas Gaur, Sarika Jain, Valerie L. Shalin, and Amit P. Sheth. 2025. "A Cross Attention Approach to Diagnostic Explainability Using Clinical Practice Guidelines for Depression." IEEE Journal of Biomedical and Health Informatics 29 (2): 1333–1342.

De Duro, Edoardo Sebastiano, Riccardo Improta, and Massimo Stella. 2025. "Introducing CounseLLMe: A dataset of simulated mental health dialogues for comparing LLMs like Haiku, LLaMAntino and ChatGPT against humans." Emerging Trends in Drugs, Addictions, and Health 5: 100170.

Ding, Zhanyi, Zhongyan Wang, Yeyubei Zhang, Yuchen Cao, Yunchong Liu, Xiaorui Shen, Yexin Tian, and Jianglai Dai. 2025. "Trade-offs between machine learning and deep learning for mental illness detection on social media." Scientific Reports 15 (1): 14497.

Farruque, Nawshad, Randy Goebel, Sudhakar Sivapalan, and Osmar R Za¨ıane. 2024. "Depression symptoms modelling from social media text: an LLM driven semi-supervised learning approach." Language Resources and Evaluation 58 (3): 1013–1041.

Ferrario, Andrea, Jana Sedlakova, and Manuel Trachsel. 2024. "The Role of Humanization and Robustness of Large Language Models in Conversational Artificial

Intelligence for Individuals With Depression: A Critical Analysis." JMIR Mental Health 11: e56569–e56569. http://dx.doi.org/10.2196/56569.

Gaber, Farieda, Maqsood Shaik, Fabio Allega, Agnes Julia Bilecz, Felix Busch, Kelsey Goon, Vedran Franke, and Altuna Akalin. 2025. "Evaluating large language model workflows in clinical decision support for triage and referral and diagnosis." npj Digital Medicine 8 (1): 263.

G´omez, Catalina, Junjie Yin, Chien-Ming Huang, and Mathias Unberath. 2024. "How large language model-powered conversational agents influence decision making in domestic medical triage contexts." Frontiers in Computer Science 6: 1427463.

Grimme, Britta, Janina Pohl, Hendrik Winkelmann, Lucas Stampe, and Christian Grimme. 2023. "Lost in transformation: rediscovering llm-generated campaigns in social media." In Multidisciplinary International Symposium on Disinformation in Open Online Media, 72– 87. Springer.

Guo, Qiming, Jinwen Tang, Wenbo Sun, Haoteng Tang, Yi Shang, and Wenlu Wang. 2024a. "Soullmate: An adaptive llm-driven system for advanced mental health support and assessment, based on a systematic application survey." arXiv preprint arXiv:2410.11859.

Guo, Yuting, Anthony Ovadje, Mohammed Ali Al-Garadi, and Abeed Sarker. 2024b. "Evaluating large language models for health-related text classification tasks with public social media data." Journal of the American Medical Informatics Association 31 (10): 2181–2189. Guo, Zhijun, Alvina Lai, Johan H Thygesen, Joseph Farrington, Thomas Keen, Kezhi Li, et al. 2024c. "Large language models for mental health applications: systematic review." JMIR mental health 11 (1): e57400.

Hermann, Karl Moritz, Tomas Kocisky, Edward Grefenstette, Lasse Espeholt, Will Kay, Mustafa Suleyman, and Phil Blunsom. 2015. "Teaching machines to read and comprehend." Advances in neural information processing systems 28.

Hua, Yining, Fenglin Liu, Kailai Yang, Zehan Li, Hongbin Na, Yi-han Sheu, Peilin Zhou, et al. 2025. "Large language models in mental health care: a scoping review." Current Treatment Options in Psychiatry 12 (1): 1–18.

Husseini Orabi, Ahmed, Prasadith Buddhitha, Mahmoud Husseini Orabi, and Diana Inkpen. 2018. "Deep Learning for Depression Detection of Twitter Users." In Proceedings of the Fifth Workshop on Computational Linguistics and Clinical Psychology: From Keyboard to Clinic, edited by Kate Loveys, Kate Niederhoffer, Emily Prud'hommeaux, Rebecca Resnik, and Philip Resnik, New Orleans, LA, Jun., 88–97. Association for Computational Linguistics. https://aclanthology.org/W18-0609/.

Ji, Shaoxiong, Tianlin Zhang, Luna Ansari, Jie Fu, Prayag Tiwari, and Erik Cambria. 2022. "MentalBERT: Publicly Available Pretrained Language Models for Mental Healthcare." In Proceedings of the Thirteenth Language Resources and Evaluation Conference, edited by Nicoletta Calzolari, Fr´ed´eric B´echet, Philippe Blache, Khalid Choukri, Christopher Cieri, Thierry Declerck, Sara Goggi, Hitoshi Isahara, Bente Maegaard, Joseph Mariani, H´el`ene Mazo, Jan Odijk, and Stelios Piperidis, Marseille, France, Jun., 7184–7190. European Language Resources Association.

https://aclanthology.org/2022.lrec-1.778/.

Jiang, Zifan, Salman Seyedi, Emily Griner, Ahmed Abbasi, Ali Bahrami Rad, Hyeokhyen Kwon, Robert O. Cotes, and Gari D. Clifford. 2023. "Multimodal mental health assessment with remote interviews using facial, vocal, linguistic, and cardiovascular patterns." http: //dx.doi.org/10.1101/2023.09.11.23295212.

Jiang, Zifan, Salman Seyedi, Emily Griner, Ahmed Abbasi, Ali Bahrami Rad, Hyeokhyen Kwon, Robert O. Cotes, and Gari D. Clifford. 2024. "Multimodal Mental Health Digital Biomarker Analysis From Remote Interviews Using Facial, Vocal, Linguistic, and Cardiovascular Patterns." IEEE Journal of Biomedical and Health Informatics 28 (3): 1680–1691. http://dx.doi.org/10.1109/JBHI.2024.3352075.

Jin, Nani, Renjia Ye, and Peng Li. 2025. "Diagnosis of depression based on facial multimodal data." Frontiers in Psychiatry 16. http://dx.doi.org/10.3389/fpsyt.2025.1508772. Jin, Yu, Jiayi Liu, Pan Li, Baosen Wang, Yangxinyu Yan, Huilin Zhang, Chenhao Ni, et al. 2025. "The Applications of Large Language Models in Mental Health: Scoping Review." Journal of Medical Internet Research 27: e69284.

Kim, Jiyeong, Kimberly G. Leonte, Michael L. Chen, John B. Torous, Eleni Linos, Anthony Pinto, and Carolyn I. Rodriguez. 2024. "Large language models outperform mental and medical health care professionals in identifying obsessive-compulsive disorder." npj Digital Medicine 7 (1). http://dx.doi.org/10.1038/s41746-024-01181-x.

Kim, Samuel, Oghenemaro Imieye, and Yunting Yin. 2025. "Interpretable Depression Detection from Social Media Text Using LLM-Derived Embeddings." arXiv preprint arXiv:2506.06616.

La Cava, Lucio, and Andrea Tagarelli. 2025. "Safeguarding Decentralized Social Media: LLM Agents for Automating Community Rule Compliance." Online Social Networks and Media 48: 100319. https://www.sciencedirect.com/science/article/pii/S2468696425000205.

Lachmar, E Megan, Andrea K Wittenborn, Katherine W Bogen, and Heather L McCauley. 2017. "# MyDepressionLooksLike: examining public discourse about depression on Twitter." JMIR mental health 4 (4): e8141.

Lan, Xiaochong, Yiming Cheng, Li Sheng, Chen Gao, and Yong Li. 2024. "Depression detection on social media with large language models." arXiv preprint arXiv:2403.10750.

Lawrence, Hannah R, Renee A Schneider, Susan B Rubin, Maja J Matari´c, Daniel J McDuff, and Megan Jones Bell. 2024b. "The Opportunities and Risks of Large Language Models in Mental Health." JMIR Mental Health 11: e59479–e59479. http://dx.doi.org/10.2196/59479.

Lawson McLean, Aaron, Yonghui Wu, Anna C Lawson McLean, and Vagelis Hristidis. 2024."Large language models as decision aids in neuro-oncology: a review of shared decisionmaking applications." Journal of Cancer Research and Clinical Oncology 150 (3): 139. Lewis, Mike, Yinhan Liu, Naman Goyal, Marjan Ghazvininejad, Abdelrahman Mohamed, Omer Levy, Veselin Stoyanov, and Luke Zettlemoyer. 2020. "BART:

Denoising Sequence-toSequence Pre-training for Natural Language Generation, Translation, and Comprehension." In Proceedings of the 58th Annual Meeting of the Association for Computational Linguistics, edited by Dan Jurafsky, Joyce Chai, Natalie Schluter, and Joel Tetreault, Online, Jul., 7871–7880. Association for Computational Linguistics. https://aclanthology.org/2020. acl-main.703/.

Gliwa, B., Mochol, I., Biesek, M., and Wawer, A. 2019. "SAMSum Corpus: A human-annotated dialogue dataset for abstractive summarization." In Proceedings of the 2nd Workshop on New Frontiers in Summarization, pp. 70–79, Hong Kong, China. Association for Computational Linguistics. Available at: https://aclanthology.org/D19-5409/DOI: 10.18653/v1/D19-5409

Li, Han, Renwen Zhang, Yi-Chieh Lee, Robert E Kraut, and David C Mohr. 2023. "Systematic review and meta-analysis of AI-based conversational agents for promoting mental health and well-being." NPJ Digital Medicine 6 (1): 236.

Li, Songtao, and Hao Tang. 2024. "Multimodal alignment and fusion: A survey." arXiv preprint. arXiv:2411.17040.

Li, Yan Chak, Linhua Wang, Jeffrey N. Law, T. M. Murali, and Gaurav Pandey. 2020. "Integrating multimodal data through interpretable heterogeneous ensembles." http://dx.doi. org/10.1101/2020.05.29.123497.

Lin, Chenhao, Pengwei Hu, Hui Su, Shaochun Li, Jing Mei, Jie Zhou, and Henry Leung. 2020. "SenseMood: Depression Detection on Social Media." In Proceedings of the 2020 International Conference on Multimedia Retrieval, ICMR '20, New York, NY, USA, 407–411. Association for Computing Machinery. https://doi.org/10.1145/3372278.3391932.

Louie, Ryan, Ananjan Nandi, William Fang, Cheng Chang, Emma Brunskill, and Diyi Yang. 2024. "Roleplay-doh: Enabling Domain-Experts to Create LLM-simulated Patients via Eliciting and Adhering to Principles." In Proceedings of the 2024 Conference on Empirical Methods in Natural Language Processing, edited by Yaser Al-Onaizan, Mohit Bansal, and Yun-Nung Chen, Miami, Florida, USA, Nov., 10570–10603. Association for Computational Linguistics. https://aclanthology.org/2024.emnlp-main.591/.

Lozoya, Daniel Cabrera, Mike Conway, Edoardo Sebastiano De Duro, and Simon D'Alfonso. 2025. "Leveraging Large Language Models for Simulated Psychotherapy Client Interactions: Development and Usability Study of Client101." JMIR Medical Education 11 (1): e68056. Ma, Zilin, Yiyang Mei, and Zhaoyuan Su. 2024. "Understanding the benefits and challenges of using large language model-based conversational agents for mental well-being support." In AMIA Annual Symposium Proceedings, Vol. 2023, 1105.

Malhotra, Anshu, and Rajni Jindal. 2022. "Deep learning techniques for suicide and depression detection from online social media: A scoping review." Applied Soft Computing 130: 109713. Masanneck, Lars, Linea Schmidt, Antonia Seifert, Tristan K¨olsche, Niklas Huntemann, Robin Jansen, Mohammed Mehsin, et al. 2024. "Triage performance across large language models, ChatGPT, and untrained doctors in emergency medicine: comparative study." Journal of medical Internet research 26:

e53297.

Mesk´o, Bertalan. 2023. “Prompt Engineering as an Important Emerging Skill for Medical Professionals: Tutorial.” Journal of Medical Internet Research 25: e50638. http://dx.doi. org/10.2196/50638.

Naseem, Usman, Adam G. Dunn, Jinman Kim, and Matloob Khushi. 2022. “Early Identification of Depression Severity Levels on Reddit Using Ordinal Classification.” In Proceedings of the ACM Web Conference 2022, WWW ’22, New York, NY, USA, 2563–2572. Association for Computing Machinery. https://doi.org/10.1145/3485447.3512128.

Nayak, Chhaya, Sachin Patel, Amruta Mahajan, and Maya Rathore. 2025. “Optimizing Mental Health Diagnostics with Hybrid Deep Learning and Multimodal Data Fusion.” International Journal of Environmental Sciences 11 (5s): 13–27. http://dx.doi.org/10.64252/sfqnv852.

Ni, Yang, and Fanli Jia. 2025. “A scoping review of AI-Driven digital interventions in mental health care: mapping applications across screening, support, monitoring, prevention, and clinical education.” In Healthcare, Vol. 13, 1205. MDPI.

Nikmehr, Golnaz, Aritz Bilbao-Jayo, Aron Henriksson, and Aitor Almeida. 2025. “Detecting Suicidal Ideation on Social Media Using Large Language Models with Zero-Shot Prompting.” In 11th International Conference on Information and Communication Technologies for Ageing Well and e-Health ICT4AWE, Porto, Portugal, 2025, 259–267. Science and Technology Publications, Lda.

Obradovich, Nick, Sahib S Khalsa, Waqas U Khan, Jina Suh, Roy H Perlis, Olusola Ajilore, and Martin P Paulus. 2024. “Opportunities and risks of large language models in psychiatry.” NPP—Digital Psychiatry and Neuroscience 2 (1): 8.

Omar, Mahmud, and Inbar Levkovich. 2025. “Exploring the efficacy and potential of large language models for depression: A systematic review.” Journal of Affective Disorders 371: 234–244.

Oniani, David, Premkumar Chandrasekar, Sonish Sivarajkumar, and Yanshan Wang. 2023. “Few-Shot Learning for Clinical Natural Language Processing Using Siamese Neural Networks: Algorithm Development and Validation Study.” JMIR AI 2: e44293. http://dx. doi.org/10.2196/44293.

Pillai, Arvind, Dimitris Spathis, Subigya Nepal, Amanda C. Collins, Daniel M Mackin, Michael V. Heinz, Tess Z Griffin, Nicholas C. Jacobson, and Andrew Campbell. 2025. “Beyond Prompting: Time2Lang Bridging Time-Series Foundation Models and Large Language Models for Health Sensing.” In Proceedings of the sixth Conference on Health, Inference, and Learning, Proceedings of Machine Learning Research, 25–27 Jun, 268–288. PMLR.

Poria, Soujanya, Erik Cambria, Devamanyu Hazarika, Navonil Mazumder, Amir Zadeh, and Louis-Philippe Morency. 2017. “Multi-level Multiple Attentions for Contextual Multimodal Sentiment Analysis.” In 2017 IEEE International Conference on Data Mining (ICDM), Nov. IEEE. http://dx.doi.org/10.1109/ICDM.2017.134.

Priyadarshana, Y. H. P. P., Ashala Senanayake, Zilu Liang, and Ian Piumarta. 2024. "Prompt engineering for digital mental health: a short review." Frontiers in Digital Health 6. http: //dx.doi.org/10.3389/fdgth.2024.1410947.

Raffel, Colin, Noam Shazeer, Adam Roberts, Katherine Lee, Sharan Narang, Michael Matena, Yanqi Zhou, Wei Li, and Peter J Liu. 2020. "Exploring the limits of transfer learning with a unified text-to-text transformer." Journal of machine learning research 21 (140): 1–67.

Rahaman, Md Abdur, Jiayu Chen, Zening Fu, Noah Lewis, Armin Iraji, Theo G. M. van Erp, and Vince D. Calhoun. 2022. "Deep multimodal predictome for studying mental disorders." Human Brain Mapping 44 (2): 509–522. http://dx.doi.org/10.1002/hbm.26077.

Rollwage, Max, Keno Juchems, Johanna Habicht, Ben Carrington, Tobias Hauser, and Ross Harper. 2022. "Conversational AI facilitates mental health assessments and is associated with improved recovery rates." medRxiv 2022–11.

Sadeghi, Misha, Robert Richer, Bernhard Egger, Lena Schindler-Gmelch, Lydia Helene Rupp, Farnaz Rahimi, Matthias Berking, and Bjoern M. Eskofier. 2024. "Harnessing multimodal approaches for depression detection using large language models and facial expressions." npj Mental Health Research 3 (1). http://dx.doi.org/10.1038/s44184-024-00112-8.

Sahili, Zahraa Al, Ioannis Patras, and Matthew Purver. 2024a. "Multimodal Machine Learning in Mental Health: A Survey of Data, Algorithms, and Challenges." arXiv preprint arXiv:2407.16804.

Sahu, Nilesh Kumar, Manjeet Yadav, Mudita Chaturvedi, Snehil Gupta, and Haroon R. Lone. 2025. "Leveraging Language Models for Summarizing Mental State Examinations: A Comprehensive Evaluation and Dataset Release." In Proceedings of the 31st International Conference on Computational Linguistics, edited by Owen Rambow, Leo Wanner, Marianna Apidianaki, Hend Al-Khalifa, Barbara Di Eugenio, and Steven Schockaert, Abu Dhabi, UAE, Jan., 2658–2682. Association for Computational Linguistics. https: //aclanthology.org/2025.coling-main.182/.

Scherbakov, Dmitry A, Nina C Hubig, Leslie A Lenert, Alexander V Alekseyenko, and Jihad SObeid. 2025. "Natural Language Processing and Social Determinants of Health in Mental Health Research: AI-Assisted Scoping Review." JMIR Mental Health 12: e67192–e67192. http://dx.doi.org/10.2196/67192.

Scholich, Till, Maya Barr, Shannon Wiltsey Stirman, and Shriti Raj. 2025. "A Comparison of Responses from Human Therapists and Large Language Model–Based Chatbots to Assess Therapeutic Communication: Mixed Methods Study." JMIR Mental Health 12 (1): e69709.

Shah, Shahid Munir, Syeda Anshrah Gillani, Mirza Samad Ahmed Baig, Muhammad Aamer Saleem, and Muhammad Hamzah Siddiqui. 2025. "Advancing depression detection on social media platforms through fine-tuned large language models." Online Social Networks and Media 46: 100311.

Shalu, Hrithwik, Hari Sankar CN, Akash Das, Saptarshi Majumder, Arnhav Datar,

Subin Mathew MS, Anugyan Das, Juned Kadiwala, et al. 2020. “Depression status estimation by deep learning based hybrid multi-modal fusion model.” arXiv preprint arXiv:2011.14966.

Sharma, Sunil Kumar, Ahmed Ibrahim Alutaibi, Ahmad Raza Khan, Ghanshyam G. Tejani, Fuzail Ahmad, and Seyed Jalaleddin Mousavirad. 2025. “Early detection of mental health disorders using machine learning models using behavioral and voice data analysis.” Scientific Reports 15 (1). http://dx.doi.org/10.1038/s41598-025-00386-8.

Shen, Guangyao, Jia Jia, Liqiang Nie, Fuli Feng, Cunjun Zhang, Tianrui Hu, Tat-Seng Chua, Wenwu Zhu, et al. 2017. “Depression detection via harvesting social media: A multimodal dictionary learning solution.” In IJCAI, Vol. 2017, 3838–3844.

Singhal, Karan, Tao Tu, Juraj Gottweis, Rory Sayres, Ellery Wulczyn, Mohamed Amin, Le Hou, et al. 2025. “Toward expert-level medical question answering with large language models.” Nature Medicine 31 (3): 943–950.

So, Jae-hee, Joonhwan Chang, Eunji Kim, Junho Na, JiYeon Choi, Jy-yong Sohn, Byung-Hoon Kim, Sang Hui Chu, et al. 2024. “Aligning large language models for enhancing psychiatric interviews through symptom delineation and summarization: pilot study.” JMIR Formative Research 8 (1): e58418.

Song, Meishu, Zijiang Yang, Andreas Triantafyllopoulos, Zixing Zhang, Zhe Nan, Muxuan Tang, Hiroki Takeuchi, et al. 2024. “Empowering Mental Health Monitoring Using a MacroMicro Personalization Framework for Multimodal-Multitask Learning: Descriptive Study.” JMIR Mental Health 11: e59512. http://dx.doi.org/10.2196/59512.

Stade, Elizabeth C, Shannon Wiltsey Stirman, Lyle H Ungar, Cody L Boland, H Andrew Schwartz, David B Yaden, Jo˜ao Sedoc, Robert J DeRubeis, Robb Willer, and Johannes C Eichstaedt. 2024. “Large language models could change the future of behavioral healthcare: a proposal for responsible development and evaluation.” NPJ Mental Health Research 3 (1): 12.

Stahlschmidt, S¨oren Richard, Benjamin Ulfenborg, and Jane Synnergren. 2022. “Multimodal deep learning for biomedical data fusion: a review.” Briefings in Bioinformatics 23 (2). http://dx.doi.org/10.1093/bib/bbab569.

Suri, Manan, Nalin Semwal, Divya Chaudhary, Ian Gorton, and Bijendra Kumar. 2023. “I don’t feel so good! Detecting Depressive Tendencies using Transformer-based Multimodal Frameworks.” In Proceedings of the 2022 5th International Conference on Machine Learning and Natural Language Processing, MLNLP ’22, New York, NY, USA, 360–365. Association for Computing Machinery. https://doi.org/10.1145/3578741.3578817.

Taylor, Niall, Andrey Kormilitzin, Isabelle Lorge, Alejo Nevado-Holgado, Andrea Cipriani, and Dan W Joyce. 2024. “Model development for bespoke large language models for digital triage assistance in mental health care.” Artificial Intelligence in Medicine 157: 102988.

Teferra, Bazen Gashaw, Argyrios Perivolaris, Wei-Ni Hsiang, Christian Kevin Sidharta, Alice Rueda, Karisa Parkington, Yuqi Wu, et al. 2025. “Leveraging large language models for automated depression screening.” PLOS Digital Health 4 (7): e0000943.

http://dx.doi. org/10.1371/journal.pdig.0000943.

Thotapalli, Samanvith, Musa Yilanli, Ian McKay, William Leever, Eric Youngstrom, Karah Harvey-Nuckles, Kimberly Lowder, et al. 2025. "Potential of ChatGPT in youth mental health emergency triage: Comparative analysis with clinicians." Psychiatry and Clinical Neurosciences Reports 4 (3): e70159.

Vajre, Vedant, Mitch Naylor, Uday Kamath, and Amarda Shehu. 2021. "PsychBERT: A Mental Health Language Model for Social Media Mental Health Behavioral Analysis." In 2021 IEEE International Conference on Bioinformatics and Biomedicine (BIBM), Dec., 1077–1082. IEEE. http://dx.doi.org/10.1109/BIBM52615.2021.9669469.

Wah, Jack Ng Kok. 2025. "Revolutionizing e-health: the transformative role of AI-poweredhybrid chatbots in healthcare solutions." Frontiers in Public Health 13: 1530799.

Wang, Jiaqi, Enze Shi, Sigang Yu, Zihao Wu, Chong Ma, Haixing Dai, Qiushi Yang, et al. 2023. "Prompt engineering for healthcare: Methodologies and applications." arXiv preprint arXiv:2304.14670.

Wang, Liying, Tanmay Bhanushali, Zhuoran Huang, Jingyi Yang, Sukriti Badami, and Lisa Hightow-Weidman. 2025. "Evaluating Generative AI in Mental Health: Systematic Review of Capabilities and Limitations." JMIR mental health 12 (1): e70014.

Wang, Ruiyi, Stephanie Milani, Jamie C. Chiu, Jiayin Zhi, Shaun M. Eack, Travis Labrum, Samuel M Murphy, et al. 2024. "PATIENT-ψ: Using Large Language Models to Simulate Patients for Training Mental Health Professionals." In Proceedings of the 2024 Conference on Empirical Methods in Natural Language Processing, edited by Yaser Al-Onaizan, Mohit Bansal, and Yun-Nung Chen, Miami, Florida, USA, Nov., 12772–12797. Association for Computational Linguistics. https://aclanthology.org/2024.emnlp-main.711/.

Wang, Xuezhi, Jason Wei, Dale Schuurmans, Quoc Le, Ed Chi, Sharan Narang, Aakanksha Chowdhery, and Denny Zhou. 2022. "Self-consistency improves chain of thought reasoning in language models." arXiv preprint arXiv:2203.11171.

Wang, Yuxi, Diana Inkpen, and Prasadith Kirinde Gamaarachchige. 2024. "Explainable Depression Detection Using Large Language Models on Social Media Data." In Proceedings of the 9th Workshop on Computational Linguistics and Clinical Psychology (CLPsych 2024), edited by Andrew Yates, Bart Desmet, Emily Prud'hommeaux, Ayah Zirikly, Steven Bedrick, Sean MacAvaney, Kfir Bar, Molly Ireland, and Yaakov Ophir, St. Julians, Malta, Mar., 108–126. Association for Computational Linguistics. https://aclanthology.org/2024.clpsych-1.8/.

Wang, Zimeng, and Yingjie Lu. 2025. "Unmasking Self-disclosure in Online Depression Communities: A Novel Framework Integrating Large Language Models and Social Network Analysis." In Wuhan International Conference on E-business, 74–88. Springer.

Wen, Bo, Raquel Norel, Julia Liu, Thaddeus Stappenbeck, Farhana Zulkernine, and Huamin Chen. 2024. "Leveraging large language models for patient engagement: The power of conversational ai in digital health." In 2024 IEEE International Conference on Digital Health (ICDH), 104–113. IEEE.

Xu, Xuhai, Bingsheng Yao, Yuanzhe Dong, Saadia Gabriel, Hong Yu, James Hendler, Marzyeh Ghassemi, Anind K. Dey, and Dakuo Wang. 2024a. “Mental-LLM: Leveraging Large Language Models for Mental Health Prediction via Online Text Data.” Proc. ACM Interact. Mob. Wearable Ubiquitous Technol. 8 (1). https://doi.org/10.1145/3643540.

Xu, Xuhai, Bingsheng Yao, Yuanzhe Dong, Saadia Gabriel, Hong Yu, James Hendler, Marzyeh Ghassemi, Anind K Dey, and Dakuo Wang. 2024b. “Mental-llm: Leveraging large language models for mental health prediction via online text data.” Proceedings of the ACM on Interactive, Mobile, Wearable and Ubiquitous Technologies 8 (1): 1–32.

Yang, Kailai, Shaoxiong Ji, Tianlin Zhang, Qianqian Xie, Ziyan Kuang, and Sophia Ananiadou. 2023. “Towards Interpretable Mental Health Analysis with Large Language Models.” In Proceedings of the 2023 Conference on Empirical Methods in Natural Language Processing, edited by Houda Bouamor, Juan Pino, and Kalika Bali, Singapore, Dec., 6056–6077. Association for Computational Linguistics. https://aclanthology.org/2023.emnlp-main.370/.

Yang, Kailai, Tianlin Zhang, Ziyan Kuang, Qianqian Xie, Jimin Huang, and Sophia Ananiadou. 2024. “MentaLLaMA: Interpretable Mental Health Analysis on Social Media with Large Language Models.” In Proceedings of the ACM Web Conference 2024, WWW ’24, New York, NY, USA, 4489–4500. Association for Computing Machinery. https://doi.org/10.1145/3589334.3648137.

Yazdavar, Amir Hossein, Mohammad Saeid Mahdavinejad, Goonmeet Bajaj, William Romine, Amit Sheth, Amir Hassan Monadjemi, Krishnaprasad Thirunarayan, et al. 2020a. “Multimodal mental health analysis in social media.” PLOS ONE 15 (4): e0226248. http: //dx.doi.org/10.1371/journal.pone.0226248.

Yu, H, and Stephen McGuinness. 2024. “An experimental study of integrating fine-tuned LLMs and prompts for enhancing mental health support chatbot system.” Journal of Medical Artificial Intelligence 1–16.

Zaghir, Jamil, Marco Naguib, Mina Bjelogrlic, Aur´elie N´ev´eol, Xavier Tannier, and Christian Lovis. 2024. “Prompt Engineering Paradigms for Medical Applications: Scoping Review.” Journal of Medical Internet Research 26: e60501. http://dx.doi.org/10.2196/60501.

Zhang, Guanghua, Guangping Zhuo, Yang Yang, Guohua Xu, Shukui Ma, Hao Liu, and Zhiyong Ren. 2025. “Sentence-level multi-modal feature learning for depression recognition.” Frontiers in Psychiatry 16. http://dx.doi.org/10.3389/fpsyt.2025.1439577.

Zhang, Jingqing, Yao Zhao, Mohammad Saleh, and Peter Liu. 2020. “Pegasus: Pre-training with extracted gap-sentences for abstractive summarization.” In International conference on machine learning, 11328–11339. PMLR.

Zhang, Tianlin, Kailai Yang, Hassan Alhuzali, Boyang Liu, and Sophia Ananiadou. 2023. “PHQ-aware depressive symptoms identification with similarity contrastive learning on social media.” Information Processing & Management 60 (5): 103417.

Zhang, Tianlin, Kailai Yang, and Sophia Ananiadou. 2023. “Sentiment-guided Transformer with Severity-aware Contrastive Learning for Depression Detection on Social Media.” In Proceedings of the 22nd Workshop on Biomedical Natural Language Processing and BioNLP Shared Tasks, edited by Dina Demner-fushman, Sophia Ananiadou, and Kevin Cohen, Toronto, Canada, Jul., 114–126. Association for Computational Linguistics. https: //aclanthology.org/2023.bionlp-1.9/.

Zhang, Yizhou, Karishma Sharma, Lun Du, and Yan Liu. 2024a. “Toward Mitigating Misinformation and Social Media Manipulation in LLM Era.” In Companion Proceedings of the ACM Web Conference 2024, WWW ’24, New York, NY, USA, 1302–1305. Association for Computing Machinery. https://doi.org/10.1145/3589335.3641256.

Zhang, Zhenwei, Shengming Zhang, Dong Ni, Zhaoguo Wei, Kongjun Yang, Shan Jin, Gan Huang, et al. 2024b. “Multimodal Sensing for Depression Risk Detection: Integrating Audio, Video, and Text Data.” Sensors, 24 (12): 3714. http://dx.doi.org/10.3390/s24123714. Zhao, Chuqing, and Yisong Chen. 2025. “LLM-powered Topic Modeling for Discovering Public Mental Health Trends in Social Media.” In IFIP International Conference on Artificial Intelligence Applications and Innovations, 119–132. Springer.

Zheng, Tong, Yanrong Guo, and Richang Hong. 2024. “Cascade Large Language Model via InContext Learning for Depression Detection on Chinese Social Media.” In Chinese Conference on Pattern Recognition and Computer Vision (PRCV), 353–366. Springer.